\documentclass[pdflatex, sn-mathphys-num]{sn-jnl}

\usepackage{amsmath,amssymb,amsfonts}%
\usepackage{amsthm}%
\usepackage{mathrsfs}%
\usepackage[title]{appendix}%
\usepackage{xcolor}%
\usepackage{textcomp}%
\usepackage{manyfoot}%
\usepackage{booktabs}%
\usepackage{algorithm}%
\usepackage{algorithmicx}%
\usepackage{algpseudocode}%
\usepackage{listings}%

\usepackage{fancyhdr}
\usepackage{graphicx}
\usepackage{multirow}
\usepackage{amscd}
\usepackage{physics}
\usepackage{rotating} 

\usepackage{longtable}
\usepackage{bbding}

\usepackage{latexsym}
\usepackage[small]{caption}

\usepackage{enumitem}

\usepackage[switch]{lineno}

\usepackage{tikz}
\usetikzlibrary{arrows,decorations.pathmorphing,backgrounds,positioning,fit,petri}
\usepackage{subcaption}
\usetikzlibrary{fit}
\usetikzlibrary{tikzmark}
\usepackage{lscape}

\usepackage{pgfplotstable}
\usepackage{array}
\usepackage{rotating}
\usepackage{placeins}
\usepackage{float}
\usepackage{url}
\usepackage[acronym]{glossaries}
\usepackage{color}
\usepackage{bm}
\usepackage{tabularx}

\usepackage{fancyvrb}

\makeglossaries
\newacronym{ai}{AI}{Artificial Intelligence}
\newacronym{pop}{POP}{Partial-Order Plan}
\newacronym{pddl}{PDDL}{Planning Domain Definition Language}
\newacronym{cg}{CG}{Causal Graph}
\newacronym{dtg}{DTG}{Domain Transition Graph}
\newacronym{fdr}{FDR}{Finite Domain Representation}
\newacronym{ipc}{IPC}{International Planning Competitions}
\newacronym{htn}{HTN}{Hierarchical Task Network}
\newacronym{pocl}{POCL}{Partial-Order Causal Link}
\newacronym{eog}{EOG}{Explanation-based Order Generalization}
\newacronym{mr}{MR}{Minimum Reordering}
\newacronym{mrr}{MRR}{Minimum Reinstantiated Reordering}
\newacronym{maxsat}{MaxSAT}{Maximum Satisfiability}
\newacronym{bd}{BD}{Block Deordering}
\newacronym{bdpo}{BDPO}{Block Decomposed Partial-Order}
\newacronym{mbd}{MBD}{Modified Block Deordering}
\newacronym{strips}{STRIPS}{Stanford Research Institute Problem Solver}
\newacronym{vhpop}{VHPOP}{Versatile Heuristic Partial Order Planner}
\newacronym{adl}{ADL}{Action Description Language}
\newacronym{cibs}{CIBS}{}
\newacronym{pbd}{PBD}{Parallel Block Decomposed}

\theoremstyle{thmstyleone}%
\newtheorem{theorem}{Theorem}
\theoremstyle{thmstyletwo}%
\newtheorem{example}{Example}%

\theoremstyle{thmstylethree}%
\newtheorem{definition}{Definition}%

\begin{document}

\title[Improving Execution Concurrency in Partial-Order Plans via Block-Substitution]{Improving Execution Concurrency in Partial-Order Plans via Block-Substitution}


\author*[1,2]{\fnm{Sabah Binte} \sur{Noor}}\email{sabah@duet.ac.bd}

\author[2,3]{\fnm{Fazlul Hasan} \sur{Siddiqui}}\email{siddiqui@duet.ac.bd}
\equalcont{These authors contributed equally to this work.}

\affil*[1]{\orgdiv{Department of Computer Science and Engineering}, \orgname{Dhaka University of Engineering \& Technoloy}, \orgaddress{\city{Gazipur}, \postcode{1700}, \country{Bangladesh}}}

\abstract{
Partial-order plans in AI planning facilitate execution flexibility and several other tasks, such as plan reuse, modification, and decomposition, due to their less constrained nature. A \acrfull*{pop} specifies partial-order over actions, providing the flexibility of executing unordered actions in different sequences. This flexibility can be further extended by enabling parallel execution of actions in the POP  to reduce its overall execution time. While extensive studies exist on improving the flexibility of a POP by optimizing its action orderings through plan deordering and reordering, there has been limited focus on the flexibility of executing actions concurrently in a plan. Flexibility of executing actions concurrently, referred to as concurrency, in a POP can be achieved by incorporating action non-concurrency constraints, specifying which actions can not be executed in parallel. This work establishes the necessary and sufficient conditions for non-concurrency constraints between two actions or two subplans with respect to a planning task. We also introduce an algorithm to improve a plan's concurrency by optimizing resource utilization through substitutions of the plan's subplans with respect to the corresponding planning task. Our algorithm employs block deordering that eliminates orderings in a POP by encapsulating coherent actions in blocks, and then exploits blocks as candidate subplans for substitutions. Experiments over the benchmark problems from International Planning Competitions (IPC) exhibit considerable improvement in plan concurrency.}

\maketitle

\newtheorem{rle}{Rule}
\newenvironment{proofsketch}{%
  \renewcommand{\proofname}{Proof Sketch}\proof}{\endproof}
\newcommand{\atom}[2]{$\langle\textit{{#1}}, \textit{#2}\rangle$}
\newcommand{\liftat}[1]{$lift$-${#1}$-$at$}
\newcommand{\boarded}[1]{$boarded$-${#1}$-$on$}
\newcommand{\passenger}[1]{$passenger$-${#1}$-$at$}
\newcommand{\dom}[1]{$D_{\text{#1}}$}
\algnewcommand{\TRUE}{\textbf{true}}
\algnewcommand{\FALSE}{\textbf{false}}
\newcommand{\flex}{$\mathit{flex}$}
\newcommand{\cflex}{$\mathit{cflex}$}

  \tikzstyle{arrow} = [thick,->,>=stealth]
\tikzstyle{process} = [rectangle,rounded corners, thick, minimum width=.6cm, minimum height=.5cm, text centered, draw=black]
\tikzstyle{block} = [draw, inner xsep= 8pt, color=black!80, fill=black!7, xshift=-2mm]
\tikzstyle{block2} = [draw, inner xsep= 8pt, color=black!80]

\section{Introduction}
AI planning automates reasoning about a plan (i.e., a sequence of actions), specifically the reasoning about formulating a plan to achieve a desired goal from a given situation \cite{lipovetzky2019introduction}. Plans are primarily categorized into sequential plans and partial-order plans (POP). A sequential plan specifies a set of actions in a strict sequential order, whereas a partial-order plan allows two actions to be executed in any order. Thus, partial-order plans are less constrained in terms of action ordering, and provide an agent the flexibility to select from various actions during execution. 
Two unordered actions in a POP can be executed in either order without affecting the plan's validity, but the POP does not specify whether the plan will remain valid if those actions are executed in parallel.
In earlier works, researchers defined parallelism based on the notion of interference \cite{RINTANEN20061031}. The parallel execution of a set of actions is possible if the actions do not interfere. These actions are referred to as \textit{concurrent (or parallel)} actions, and the plan that allows such actions is called a parallel plan. Notably, in the classical setting, concurrency and executability in any order coincide for actions in POPs \cite{RINTANEN20061031}, meaning that unordered actions in a POP are also concurrent. Therefore, to improve concurrency in a POP, we need to minimize its action orderings. Optimizing orderings in a POP has been studied through \emph{plan deordering}, which eliminates unnecessary action orderings, and
plan reordering, which modifies action orderings arbitrarily. Block deordering \cite{siddiqui_patrik_2012} can further remove action orderings from a POP by encapsulating action sets (i.e., subplans) in blocks, resulting in a hierarchical structured plan named Block Decomposed Partial-Order (BDPO) plan. Unlike traditional POP, two unordered blocks of actions (i.e., subplans) are not always concurrent. This scenario will be further illustrated in Example \ref{ex:example1}. In this work, we formalize the semantics for enabling parallel execution in BDPO plans, and introduce algorithms to improve their concurrency by parallelizing unordered, yet non-concurrent, blocks within these plans.

In this study, we adopt the notion of parallel plans introduced by Bäckström \cite{backstorm1994}. Bäckström formalizes parallel plans by incorporating additional constraints, referred to as \emph{non-concurrency constraints}, in partial-order plans. These non-concurrency constraints specify which actions can not be executed in parallel, thereby allowing concurrent execution of unordered actions with no such constraint in POPs. 
This work introduces a novel algorithm aimed at minimizing non-concurrency action constraints to enhance the parallel execution flexibility of plans. We employ post-processing techniques that reduce both ordering and non-concurrency action constraints in a plan. Specifically, our approach leverages deordering techniques, such as \acrfull*{eog} \cite{kk, Veloso2002} and block deordering \cite{siddiqui_patrik_2012}, to remove action orderings. \acrshort*{eog} efficiently transforms a sequential plan into a POP in polynomial time by preserving only the necessary orderings. Subsequently, block deordering further reduces ordering constraints by encapsulating coherent action sets (i.e., subplans) into blocks, producing a \acrfull*{bdpo} plan. We then exploit these blocks as candidate subplans for substitutions to minimize non-concurrency constraints in the corresponding BDPO plan. In previous works \cite{siddiqui_patrik_2012, maxsat}, the flexibility of executing actions in different orders is calculated by the ratio of unordered action pairs to the total number of action pairs and is referred to as \flex{}. Similarly, we estimate the concurrency of a plan, denoted as \cflex{}, by the ratio of concurrent action pairs to the total number of action pairs.

We implement our proposed methods on the code base of the fast downward planning system \cite{fd}. Fast Downward is a classical planning system that handles general deterministic planning problems, including advanced features such as ADL conditions, effects, and derived predicates (axioms) \cite{lipovetzky2019introduction}. We experiment our strategies with the PDDL benchmark problems from International Planning Competitions (IPC) \cite{icaps} and perform a comparative analysis.

\subsection{An Illustrative Example}

We illustrate how block deordering and block-substitution can improve plan concurrency via block-substitution using a simple plan from \emph{Elevator} domain in Example \ref{ex:example1}. The tasks of this domain are to transfer passengers from one floor to another using lifts.  We also use plans from the same domain to clarify different core concepts across paper.

\begin{example}\label{ex:example1}
\small
\begin{figure}[!bp]
\centering
    \begin{subfigure}[t]{.4\textwidth}
        \begin{tikzpicture}[node distance = .7cm]
            \node (op1)[rectangle, text width = 4.5cm]{\verb| 1 (move_down e1 n3 n2)|};
            \node (op2)[rectangle, text width = 4.5cm, below of = op1]{\verb| 2 (board p1 n2 e1)|};
            \node (op3)[rectangle, text width = 4.5cm, below of = op2]{\verb| 3 (board p2 n2 e1)|};
            \node (op4)[rectangle, text width = 4.5cm, below of = op3]{\verb| 4 (move_up e1 n2 n3)|};
            \node (op5)[rectangle, text width = 4.5cm, below of = op4]{\verb| 5 (leave p1 n3 e1)|};
            \node (op6)[rectangle, text width = 4.5cm, below of = op5]{\verb| 6 (leave p2 n3 e1)|};
            \node (op7)[rectangle, text width = 4.5cm, below of = op6]{\verb| 7 (move_down e1 n3 n2)|};
            \node (op8)[rectangle, text width = 4.5cm, below of = op7]{\verb| 8 (move_down e1 n2 n1)|};
            \node (op9)[rectangle, text width = 4.5cm, below of = op8]{\verb| 9 (board p3 n1 e1)|};
            \node (op10)[rectangle, text width = 4.5cm, below of = op9]{\verb|10 (move_up e1 n1 n2)|};
            \node (op11)[rectangle, text width = 4.5cm, below of = op10]{\verb|11 (leave p3 n2 e1)|};
        \end{tikzpicture}
        \caption{}
    \end{subfigure}
    \centering
    \begin{subfigure}[t] {.58\textwidth}
        \begin{tikzpicture}[node distance = .8cm]
            \node (init)[process] {\verb|INIT|};
            \node (op1)[process, below of = init]{\verb|move_down e1 n3 n2|};
            \node (op2)[process, below of = op1, xshift = -2cm]{\verb|board p1 n2 e1|};
            \node (op3)[process, below of = op1, xshift = 2cm]{\verb|board p2 n2 e1|};
            \node (op4)[process, below of = op2,  xshift = 2cm]{\verb|move_up e1 n2 n3|};
            \node (op5)[process, below of = op4, xshift = -2cm]{\verb|leave p1 n3 e1|};
            \node (op6)[process, below of = op4, xshift = 2cm]{\verb|leave p2 n3 e1|};
            \node (op7)[process, below of = op5, xshift = 2cm]{\verb|move_down e1 n3 n2|};
            \node (op8)[process, below of = op7]{\verb|move_down e1 n2 n1|};
            \node (op9)[process, below of = op8]{\verb|board p3 n1 e1|};
            \node (op10)[process, below of = op9]{\verb|move_up e1 n1 n2|};
            \node (op11)[process, below of = op10]{\verb|leave p3 n2 e1|};
            \node (goal)[process, below of= op11] {\verb|GOAL|};
             \draw [arrow] (init) ->(op1);

            \draw [arrow] (op1) ->(op2); 
            \draw [arrow] (op1) ->(op3); 
            \draw [arrow] (op2) ->(op4); 
             \draw [arrow] (op3) ->(op4); 
            \draw [arrow] (op4) ->(op5); 
            \draw [arrow] (op4) ->(op6);
            \draw [arrow] (op5) ->(op7);
            \draw [arrow] (op6) ->(op7);  
            \draw [arrow] (op7) ->(op8); 
             \draw [arrow] (op8) ->(op9);  
            \draw [arrow] (op9) ->(op10); 
            \draw [arrow] (op10) ->(op11);
            \draw [arrow] (op11) ->(goal); 
        \end{tikzpicture}
        \caption{}
    \end{subfigure}
      \caption{(a) A sequential plan from \emph{Elevator} domain, where a lift \texttt{e1} transfers passengers \texttt{p1} and \texttt{p2} from floor \texttt{n2} to \texttt{n3}, and passenger \texttt{p3} from floor \texttt{n1} to \texttt{n2}. (b) A partial-order plan with \flex{} and \cflex{} values of 0.037, generated from the sequential plan in (a) using \acrshort*{eog}.}
    \label{fig:pop}
    
\end{figure}
 Let us consider a plan from the \emph{Elevator} domain presented in Figure \ref{fig:pop}(a).   This plan moves passenger $p1$ and $p2$  from floor $n2$ to $n3$, and passenger $p3$ from floor $n1$ to $n2$ using lift $e1$. Actions \verb|move_up| and  \verb|move_down| move a lift up and down by one floor, respectively. On the other hand, actions \verb|board| and \verb|leave| take a passenger in and out of a lift on a specific floor, respectively.

\begin{figure}[!tbp]
    \centering
    \small{
    \begin{tikzpicture}[node distance = .8cm]
        \node (init)[process] {\verb|INIT|};
        \node (op1)[process, below of = init]{\verb|move_down e1 n3 n2|};
        \node (op2)[process, below of = op1, xshift = -3.6cm, yshift=-.4cm]{board p1 n2 e1};
        \node (op3)[process, below of = op1, xshift = -1cm, yshift=-.4cm]{\verb|board p2 n2 e1|};
        \node (op4)[process, below of = op3, xshift=-1.5cm]{\verb|move_up e1 n2 n3|};
        \node (op5)[process, below of = op4, xshift = -1.4cm ]{\verb|leave p1 n3 e1|};
        \node (op6)[process, below of = op4, xshift=1.4cm]{\verb|leave p2 n3 e1|};
        \node (op7)[process, below of = op6, xshift=-1.5cm]{\verb|move_down e1 n3 n2|};
        \node (op8)[process, below of = op1, xshift =3cm,  yshift=-.4cm]{\verb|move_down e1 n2 n1|};
        \node (op9)[process, below of = op8]{\verb|board p3 n1 e1|};
        \node (op10)[process, below of = op9]{\verb|move_up e1 n1 n2|};
        \node (op11)[process, below of = op10]{\verb|leave p3 n2 e1|};
        \node (goal)[process, below of= op11,  xshift =-3cm, yshift=-.4cm] {\verb|GOAL|};
        
        \draw [arrow] (init) ->(op1);            
        \draw [arrow] (op2) ->(op4); 
        \draw [arrow] (op3) ->(op4); 
        \draw [arrow] (op4) ->(op5);
        \draw [arrow] (op4) ->(op6);
        \draw [arrow] (op5) ->(op7);
        \draw [arrow] (op6) ->(op7);  
        \draw [arrow] (op8) ->(op9); 
        \draw [arrow] (op9) ->(op10);  
        \draw [arrow] (op10) ->(op11); 

         \begin{pgfonlayer}{background}
             \node (fit1) [fit=(op2) (op3) (op4) (op5) (op6) (op7)] [block2, label={[xshift=5.5mm,]left:{$\mathbf{b_1}$}}] {};
             \node (fit2) [fit=(op8) (op9) (op10) (op11)] [block2, label={[xshift=5.5mm]left:{$\mathbf{b_2}$}}]{};
        \end{pgfonlayer}
            \draw [arrow] (fit2.south) ->(goal); 
            \draw [arrow] (fit1.south) ->(goal); 
            \draw [arrow] (op1) ->(fit1.north);
            \draw [arrow] (op1) ->(fit2.north);
            \draw[ultra thick] (fit1) -- node[yshift=.3cm]{\#}(fit2);
    \end{tikzpicture}
    }
    \caption{A block-decomposed partial-order (BDPO) plan, generated from the POP in Figure \ref{fig:pop}(b), allows the unordered blocks, $b_1$ and $b_2$, to be executed in any order but not in parallel, yielding a $flex$ of 0.47 and a $cflex$ of 0.037.}
    \label{fig:dpop}
\end{figure}

Figure \ref{fig:pop}(b) shows a partial-order plan, generated from the sequential plan in Figure \ref{fig:pop}(a)  using \acrshort*{eog}. The POP in  Figure \ref{fig:pop}(b)  has two pairs of unordered and parallel actions: i) \verb|board p1 n2 e1| and \verb|board p2 n2 e1|, and ii) \verb|leave p1 n3 e1| and \verb|leave p2 n3 e1|, yielding \flex{} and \cflex{} values of 0.037 (two unordered, concurrent pairs out of a total of 55 pairs). Block deordering eliminates orderings from this POP by constructing two blocks $b_1$ and $b_2$ over actions 2 to 7 and 8 to 11, respectively. In the resultant BDPO plan (presented in Figure \ref{fig:dpop}), blocks $b_1$ and $b_2$ have no ordering, thereby yielding a BDPO plan with a flex of 0.47 (26 unordered action pairs out of a total of 55 pairs). However, these two blocks cannot be executed in parallel as they use the same resource, lift \verb|e1|. Hence, these blocks have a non-concurrency constraint, represented using \#. Therefore, the BDPO plan still has a \cflex{} value of 0.037.

If a second lift $e2$ exists, we can enhance the concurrency of this BDPO plan by substituting blocks. Considering that the lift $e2$ is initially on floor $n1$, we can replace the block $b_2$ with the following subplan by encapsulating it in another block $b_2'$.

\begin{verbatim}
1 (board p3 n1 e2)
2 (move_up e2 n1 n2)
3 (leave p3 n2 e2)    
\end{verbatim}

\begin{figure}[!btp]
    \centering
    \small{
    \begin{tikzpicture}[node distance = .8cm]
                \node (init)[process] {\verb|INIT|};
                \node (op1)[process, below of = init]{\verb|move_down e1 n3 n2|};
                \node (op2)[process, below of = op1, xshift = -3.8cm, yshift=-.4cm]{\verb|board p1 n2 e1|};
                \node (op3)[process, below of = op1, xshift = -1cm, yshift=-.4cm]{\verb|board p2 n2 e1|};
                \node (op4)[process, below of = op3, xshift=-1.5cm]{\verb|move_up e1 n2 n3|};
                \node (op5)[process, below of = op4, xshift = -1.4cm ]{\verb|leave p1 n3 e1|};
                \node (op6)[process, below of = op4, xshift=1.4cm]{\verb|leave p2 n3 e1|};
                \node (op7)[process, below of = op6, xshift=-1.5cm]{\verb|move_down e1 n3 n2|};
                
                \node (op8)[process, below of = op1, xshift =3cm,  yshift=-.8cm]{\verb|board p3 n1 e2|};
                \node (op9)[process, below of = op8]{\verb|move_up e2 n1 n2|};
                \node (op10)[process, below of = op9]{\verb|leave p3 n2 e2|};
                \node (goal)[process, below of= op10,  xshift =-3cm, yshift=-.6cm] {\verb|GOAL|};
                
                \draw [arrow] (init) ->(op1);            
                \draw [arrow] (op2) ->(op4); 
                 \draw [arrow] (op3) ->(op4); 

                \draw [arrow] (op4) ->(op5);
                \draw [arrow] (op4) ->(op6);
                \draw [arrow] (op5) ->(op7);
                \draw [arrow] (op6) ->(op7);  
                \draw [arrow] (op8) ->(op9);

                \draw [arrow] (op9) ->(op10);  
                 
                 \begin{pgfonlayer}{background}
                     \node (fit1) [fit=(op2) (op3) (op4) (op5) (op6) (op7)] [block2, label={[xshift=5.5mm]left:{$\mathbf{b_1}$}}] {};
                     \node (fit2) [fit=(op8) (op9) (op10)] [block2, label={[xshift=5.5mm, yshift=.5cm]left:{$\mathbf{b_2'}$}}]{};
                \end{pgfonlayer}
                    \draw [arrow] (fit2.south) ->(goal); 
                    \draw [arrow] (fit1.south) ->(goal); 
                    \draw [arrow] (op1) ->(fit1.north);
                    \draw [arrow] (op1) ->(fit2.north);
            \end{tikzpicture}
    }
    \caption{ Substituting block $b_2$ in the BDPO plan (presented in Figure \ref{fig:dpop}) with the block $b_2'$  allows the unordered blocks $b_1$ and $b_2'$ to be executed in any order, increasing the plan \cflex{} from 0.037 to 0.47. }
    \label{fig:dpop2}
\end{figure}
 
This substitution allows blocks $b_1$ and $b_2'$ to be executed in parallel since these two blocks are using two different lifts (shown in Figure \ref{fig:dpop2}), thus improving the \cflex{} from 0.037 to 0.47. 
 \end{example}


\subsection{Contributions}
The following are the main contributions of this paper:
\begin{itemize}
    \item We propose necessary and sufficient conditions for non-concurrency constraints between two actions or subplans w.r.t. a planning problem.
    \item We present the concept of \acrfull*{pbd}, by presenting the necessary semantics for enabling parallelism in BDPO plans.
    \item We introduce a procedure to extend a block (i.e., subplan) incorporating its cohesive neighbors to capture  subplans responsible for different subtasks of the
    corresponding problem.
    \item We develop an algorithm for improving a plan's concurrency by substituting its subplans w.r.t. the corresponding planning task. 
    \item We introduce $cflex$, which captures a normalized measure of the number of non-concurrency constraints. We also establish the empirical connection between the minimum duration of a plan and its $cflex$ measure.
    \item Our experimental result on plans from IPC domains demonstrate our method improves \cflex{}, thereby minimum duration of considerable number of plans in some domains. This result justifies the need for an approach such as ours to compute plans with more concurrency. 
\end{itemize}

\subsection{Organization}
We start by presenting the essential background concepts and notation in Section 2. In Section 3, we introduce necessary semantics to enable parallelism in BDPO plans. Next, we discuss the conditions for non-concurrency constraints in Section 4. In Section 5, we detail our approach to enhancing concurrency in plans. Following that, we present our evaluation in Section 6, and we conclude with our remarks in Section 7.

\section{Preliminaries}
 This section explains the semantics of a planning task, partial-order planning, and parallel plan.  We also present related plan deordering techniques (e.g., \acrshort*{eog} and block deordering) and block substitution strategy \cite{noor2024improving}. Additionally, we discuss previous approaches to enabling the parallel execution of actions in a plan. 

\subsection{Planning Task}
This paper considers classical planning in \acrfull*{fdr}  \cite{fd} due to its compact and structured representation. \acrshort*{fdr} represents a planning task using a set of state variables and their corresponding values. In contrast with propositional encodings, \acrshort*{fdr} can capture the internal structure and the behavior of state variables of a planning instance through concise constructs such as Causal Graph (CG) and Domain Transition Graph (DTG). We exploit the informative structure of \acrshort*{fdr} in our methods, presented in Sections \ref{sec:conditions} and \ref{sec:sec4}. Planning tasks are commonly described using \acrfull*{pddl} \cite{lipovetzky2019introduction}, which can be automatically converted into FDR \cite{HELMERT2009503}. 

\begin{definition}
\label{def:sas}
A \textbf{planning task in FDR} is a 4-tuple $\Pi =\langle \mathcal{V}, O, s_0, s_* \rangle$ where:
\begin{itemize}
    \item $\mathcal{V}$ is a finite set of \textbf{state variables}, each with an associated finite domain $\mathcal{D}_v$. A \textbf{fact} is a pair $\langle v,d\rangle$ with $v\in \mathcal{V}$ and $d\in \mathcal{D}_v$. \\
    A \textbf{state} is a function $s$ defined on $\mathcal{V}$, where for all $v \in \mathcal{V}$, there must be $s(v)\in\mathcal{D}_v$. We often notationally treat a state as a set of facts.  A \textbf{partial state} is essentially the function $s$ but defined only on a subset of $\mathcal{V}$,  denoted as $\mathit{vars}(s)$.
    \item O is a finite set of \textbf{operators}. Each operator $o$ has an associated partial state  $pre(o)$ called its \textbf{precondition}, an associated partial state $\mathit{eff}(o)$ called its \textbf{effect}, and an associated nonnegative number   $cost(o) \in \mathbb{R}^+_0$ called its \textbf{cost}.
    \item $s_0$ is the \textbf{initial state}, 
    \item $s_*$ is a partial state representing \textbf{goal} conditions.
\end{itemize}
\end{definition}

An operator $o$ is applicable in a state $s$ iff $\mathit{pre}(o) \subseteq s$, and applying $o$ in $s$ yields another state $\hat{s} = apply(o,s)$.  In  the state $\hat{s}$, the value of $v$ is $d$ for each $\langle v, d \rangle \in \mathit{eff}(o)$. 

A plan $\pi$ is a sequence of operators $\langle o_1, o_2, \dots, o_i ,\dots, o_n \rangle$ and is valid for a planning task $\Pi$ iff,
\begin{enumerate}
    \item $\mathit{pre}(o_1)\subseteq s_0$,
    \item  $\forall i \in \{1,2,\dots,n-1\}$ $\mathit{pre}(o_{i+1}) \subseteq s_i$, where $s_i= apply(o_i,s_{i-1})$, and
    \item $s_* \subseteq s_n$, where $s_n= apply(o_n,s_{n-1})$
\end{enumerate}

For the rest of this paper, we will use the term \emph{operator} instead of \emph{action}, as FDR refers to ground actions as operators (Definition \ref{def:sas}).  An FDR operator does not explicitly state its add or delete effects. Definition \ref{def:op} specifies the facts that an operator consumes, produces, and deletes in FDR.  An operator $o$  produces a fact $\langle v, d \rangle$ if $\langle v, d \rangle$ belongs to $o$'s effect. On the other hand, $o$ deletes a fact $\langle v, d \rangle$ if $o$ changes the value of $v$ from $d$ to another value. 

\begin{definition}
\label{def:op}
  The set of facts that are consumed, produced, and deleted by an operator $o$ are denoted as \textbf{cons(o)}, \textbf{prod(o)}, and \textbf{del(o)}, respectively.
    \begin{itemize}
        \item A fact $\langle v, d \rangle\in cons(o)$ iff $\langle v, d \rangle \in \mathit{pre}(o)$.
        \item  A fact $\langle v, d \rangle\in prod(o)$ iff $\langle v, d \rangle \in \mathit{eff}(o)$.
         \item  A fact $\langle v, d \rangle \in del(o)$ iff 
            \begin{enumerate}[label=\arabic*.]
                \item  Either  $v \notin vars(\mathit{cons}(o))$ or $\mathit{cons}(o)(v) =  d$,  and 
                \item  $\mathit{eff}(o)(v) =d'$ s.t. $\ d' \in (\mathcal{D}_v\setminus \{d\})$.  
                
            \end{enumerate}
    \end{itemize}
\end{definition}

Notably, $cons(o), prod(o)$, and $del(o)$ are partial states, and $cons(o)(v)$ denotes the value of the variable $v$ in $cons(o)$. When an operator $o$ does not consume a fact with variable $v$ (i.e.,  $v \notin vars(\mathit{cons}(o))$) and sets the variable $v$ to a value other than $d$, the state in which $o$ is applied determines whether $o$ deletes \atom{v}{d} or not. Since the current state is unavailable,  Definition \ref{def:op} states that $o$ deletes $\langle v, d \rangle$ in this scenario to prevent the possibility of overlooking a potential deleter. This definition will be expanded further in the subsequent section.

We exploit \acrfull*{dtg} to capture the interdependencies and behavior of state variables and operators in FDR. The \acrfull*{dtg} for a variable $v$, denoted as $G_v$, has an edge from $d$ to $d'$ if there exists some operator that can change the value of $v$ from $d$ to $d'$. For a DTG $G_v$, we define the \emph{transition path set} $P(G_v, d, d')$ to be the set of all possible paths from $d$ to $d'$ in $G_v$.

\begin{definition}
\textbf{\acrfull*{dtg}} for a variable $v\in V$, denoted as $G_v$, is a labeled directed graph with vertex set $D_v$ and edges $(d, d')$ labeled with operator  $o \in \mathcal{O}$ if either 
\begin{enumerate}
    \item $pre(o)(v) = d$ and $\mathit{eff}(o)(v)=d'$, or
    \item $v\notin vars(pre(o))$ and $\mathit{eff}(o)(v)=d'$.
\end{enumerate}
\end{definition}

In Example \ref{example:sas}, we illustrate these concepts using a planning task from the \emph{Elevator} domain. This example is a simplified version of the one presented in Example \ref{ex:example1}. We will refer to this example throughout the paper to clarify and demonstrate various concepts.

\begin{example}\label{example:sas}
    Let us consider a planning task from \emph{Elevator} domain, where the goal is to transfer one passenger \verb|p1| from floor \verb|n2| to floor \verb|n3| and another passenger from floor \verb|n1| to \verb|n2|.  We describe this planning task using four variables $v_{e1}, v_{e2}, v_{p1}$, and $v_{p2}$  in FDR. Variables $v_{e1}$ and $v_{e2}$ represent locations of lifts \verb|e1| and \verb|e2|, respectively, each having a domain of three values specifying three floors (\verb|n1|, \verb|n2|, and \verb|n3|).  The variables $v_{p1}$ and $v_{p2}$ represent the locations of the passengers \verb|p1| and \verb|p2|, respectively, each with a domain of five values (\verb|n1|, \verb|n2|, \verb|n3|, \verb|e1|, and \verb|e2|) indicating three floors and two lifts. Table \ref{tbl:sas-example} presents the variable domains, start state, and goal conditions denoted by $\mathcal{D}, s_0$, and $s_*$, respectively.
    The preconditions and effects of some operators from this task are given in Table \ref{tbl:sas-operator2}. Figure \ref{fig:dtg1}(a)  presents the \acrshort*{dtg} for the variables $v_{p_1}$ and $v_{p2}$. This graph indicates valid transitions between different values of these variables, indicating that an operator exists to facilitate the transition.  Similarly,  Figure \ref{fig:dtg1}(b)  illustrates the \acrshort*{dtg} for the variables $v_{e1}$ and $v_{e2}$.

\begin{table}[!tbp]
    \caption{A FDR representation of the Elevator planning task, described in Example \ref{example:sas}, where variables $v_{e1}$ and $v_{e2}$ specify the floor positions of two lifts, and variables $v_{p1}$ and $v_{p2}$ state the positions of the two passengers.}
    \label{tbl:sas-example}
        \centering
        \begin{tabularx}{.8\textwidth}{l X}
        \toprule
            $\mathcal{V}$ &  \{$v_{e1}, v_{e2}, v_{p1}, v_{p2}, v_{p3}$\}\\
            \dom{$v_{e1}$} &   $\{  n1, n2, n3 \}$ \\     
            \dom{$v_{e2}$} & $\{ n1, n2, n3 \}$ \\    
            \dom{$v_{p1}$} &  $\{ n1, n2, n3, e1, e2\}$ \\
            \dom{$v_{p2}$} &  $\{n1, n2, n3, e1, e2\}$ \\
            $s_0$ &  \{\atom{$v_{e1}$}{n2}, \atom{$v_{e2}$}{n2} , \atom{$v_{p1}$}{n2}, \atom{$v_{p2}$}{n1}  \\
            $s_*$ &  \{\atom{$v_{p1}$}{n3}, \atom{$v_{p2}$}{n2}\}\\
            \bottomrule
        \end{tabularx}

\end{table}

\begin{table}[!tbp]
\caption{Preconditions and Effects of operators of an Elevator planning instance. The variables are specified in Table \ref{tbl:sas-example}. }
    \label{tbl:sas-operator2}
    \centering
    \begin{tabularx}{.8\textwidth}{X X X}
    \toprule
    Name & Precondition & Effect \\
    \midrule
        \verb|board p1 n1 e1| & \{\atom{$v_{p1}$}{n1}, \atom{$v_{e1}$}{n1}\} & \{\atom{$v_{p1}$}{e1}\} \\
        \verb|board p2 n1 e1| & \{\atom{$v_{p2}$}{n1}, \atom{$v_{e1}$}{n1}\} & \{\atom{$v_{p2}$}{e1}\} \\
        \verb|board p2 n2 e1| & \{\atom{$v_{p2}$}{n2}, \atom{$v_{e1}$}{n2}\} & \{\atom{$v_{p2}$}{e1}\} \\
        \verb|move_up e1 n1 n2| & \{\atom{$v_{e1}$}{n1}\} & \{\atom{$v_{e1}$}{n2}\} \\
        \verb|move_down e1 n2 n1| & \{\atom{$v_{e1}$}{n2}\} & \{\atom{$v_{e1}$}{n1}\} \\
        \verb|move_up e2 n2 n3| & \{\atom{$v_{e2}$}{n2}\} & \{\atom{$v_{e2}$}{n3}\} \\
        \dots & \dots & \dots \\
        \bottomrule
    \end{tabularx}
\end{table}

\begin{figure}[!tbp]
    \begin{subfigure}{.48\textwidth}
    \centering
        \begin{tikzpicture}[node distance = 2.5cm]
                \node (n1)[process] {$n_1$};
                \node (e1)[process, right of=n1] {$e1$};
                \node (n2)[process, below of=e1]{$n_2$};
                \node (e2)[process, left of=n2]{$e_2$};
                \node (n3)[process, left of=n2, yshift=1.25cm, xshift= 1.25cm]{$n_3$};
                
                \draw [arrow] (n1) ->(e1);
                \draw [arrow] (e1) ->(n1);
                \draw [arrow] (n2) ->(e1);
                \draw [arrow] (e1) ->(n2);
                
                \draw [arrow] (e2) ->(n2);
                \draw [arrow] (n2) ->(e2);
                  \draw [arrow] (e2) ->(n1);
                \draw [arrow] (n1) ->(e2);
                \draw [arrow] (n3) ->(e1);
                \draw [arrow] (e1) ->(n3);
                \draw [arrow] (e2) ->(n3);
                \draw [arrow] (n3) ->(e2);
            \end{tikzpicture}
            \caption{\acrshort*{dtg} variables $v_{p1}$ and $v_{p2}$.}
     \end{subfigure}
    \begin{subfigure}{.48\textwidth}
    \centering
        \begin{tikzpicture}[node distance = 2cm]
                \node (n)[rectangle]{};
                \node (n1)[process, above of=n, yshift=-.5cm] {$n_1$};
                \node (n2)[process, right of=n1] {$n_2$};
                \node (n3)[process, right of=n2]{$n_3$};
                
                \draw [arrow] (n1) ->(n2);
                \draw [arrow] (n2) ->(n1);
                \draw [arrow] (n2) ->(n3);
                \draw [arrow] (n3) ->(n2);
            
            \end{tikzpicture}
             \caption{\acrshort*{dtg}  for  variables $v_{e1}$ and $v_{e2}$. }
    \end{subfigure}
   
    \caption{Domain Transition Graph (DTG) for the variables in the Elevator planning task, presented in the Example \ref{example:sas}.}
    \label{fig:dtg1}
\end{figure}
\end{example}

\subsection{Partial-Order Planning}
A partial-order plan (POP) allows unordered operators to be executed in any sequence by specifying a partial order over its operators.  Though an operator can appear more than once in a POP, Definition \ref{def:pop} defines a POP w.r.t, a planning task $\Pi=\langle \mathcal{V}, O, s_0, s^*\rangle$, assuming that every plan operator is uniquely identifiable.
\begin{definition}\label{def:pop}
A \textbf{partial-order plan} is a 2-tuple $\pi_{pop} =\langle \mathcal{O}, \prec \rangle$ where:
\begin{itemize}
    \item $\mathcal{O}$ is a set of operators.
    \item $\prec$ is a set of ordering constraints over $\mathcal{O}$. An \textbf{ordering constraint} between a pair of operators, $o_i$ and $o_j$  s.t. $o_i, o_j\in \mathcal{O}$, written as $o_i \prec o_j$, states that the operator $o_i$  must be executed anytime before the operator $o_j$. 
\end{itemize}
\end{definition}
 
The ordering constraint $\prec$ is transitive, meaning that if $o_i \prec o_j$ and $o_j \prec o_k$, then $o_i \prec o_k$. An ordering $o_i \prec o_j$ is considered a \emph{basic ordering} if it is not transitively implied by other orderings in $\prec$. A \emph{linearization} of $\pi_{pop}$ is a total ordering of the operators in $\mathcal{O}$. In the following sections, we often use $\nprec$ notation, where $o_i \nprec o_j$ indicates that $o_i$ is not ordered before $o_j$.

To form a POP's ordering structure,  the \emph{producer-consumer-threat} (PCT) formalism \cite{backstrom1998} first identifies which operators produce, consume, or delete which facts, and then maps each precondition of an operator to another operator's effect by constructing operator orderings. These orderings are referred to as causal links. A threat occurs when an operator $o_t$ deletes a fact, and can be executed between two operators $o_i$ and $o_j$, where a causal link exists from  operator $o_i$  to operator $o_j$ for providing the fact. 

\begin{definition}
A \textbf{causal link} between $o_i$ and $o_j$, written as $o_i \xrightarrow{\langle v, d \rangle} o_j$, specifies that  $o_i \prec o_j$ and the operator $o_i$  provides a fact \atom{v}{d} to the operator $o_j$ where $\langle v, d \rangle\in (prod(o_i) \cap cons(o_j))$. 
\end{definition}

\begin{definition}
    A \textbf{threat} represents a conflict between an effect of an operator $o_t$ and a causal link $o_i \xrightarrow{\langle v, d \rangle} o_j$. Operator $o_t$ \textbf{threatens}  $o_i \xrightarrow{\langle v, d \rangle} o_j$ if $o_t$ deletes $\langle v, d \rangle$ and can be ordered between $o_i$ and $o_j$.
\end{definition}

 A threat between an operator $o_t$ and a causal link $o_i \xrightarrow{\langle v, d \rangle} o_j$ can be resolved either by a \textbf{promotion}, adding an ordering constraint $o_t \prec o_i$, or by a \textbf{demotion}, adding an ordering constraint $o_j \prec o_t$. A POP $\pi$ is valid iff every operator precondition is supported by a causal link with no threat \cite{weld_1994}. 

  Siddiqui and Haslum \cite{siddiqui_patrik_2012} introduce three labels, namely  $PC, CD$, and $DP$, to annotate the ordering constraints in a POP, defined as follows. 

 \begin{definition} Let $\pi_{pop} =\langle \mathcal{O}, \prec \rangle$ be a partial-order plan, and $Re(o_i \prec o_j)$ be the set of \emph{ordering reasons} between two operators $o_i$ and $o_j$. $o_i \prec o_j$ can be formed due to three types of reasons, labeled as \textbf{PC}, \textbf{CD}, and \textbf{DP}.
     \begin{itemize}
    \item PC(\atom{v}{d}): Producer-consumer of a fact \atom{v}{d}, $PC(\langle v, d \rangle) \in Re(o_i \prec o_j)$, occurs when  $o_i$ produces  $\langle v, d \rangle$ and $o_j$ consumes $\langle v, d \rangle$. Multiple operators may produce $\langle v, d \rangle$, and similarly, multiple operators can also consume $\langle v, d \rangle$.  $PC$ assigns one operator $o_i$ to achieve $\langle v, d \rangle$ for the operator $o_j$.
    
    \item CD(\atom{v}{d}): Consumer-deleter of a fact \atom{v}{d}, $CD(\langle v, d \rangle) \in Re(o_i \prec o_j)$, occurs when operator $o_i$ consumes the fact   $\langle v, d \rangle$ and $o_j$ also  deletes $\langle v, d \rangle$.

    \item DP(\atom{v}{d}):  Deleter-producer of a fact  $\langle v, d \rangle$, $DP(\langle v, d \rangle) \in Re(o_i \prec o_j)$,  occurs when an operator $o_i$
    deletes $\langle v, d \rangle$ and there is at least one causal link $o_j \xrightarrow{\langle v, d \rangle} o_k$ for some operator $o_k\in \mathcal{O}$. 
\end{itemize}
\label{def:pc-cd-dp}
 \end{definition}

 $PC$ represents the causal links, whereas $CD$ and $DP$ signify the demotion and promotion orderings, respectively, in a POP. These labels help to identify and trace the reasoning behind orderings within a POP.

\begin{example}\label{example:pop}
The following is a sequential plan for the planning task presented in Example \ref{example:sas}.  Figure \ref{fig:ordering-reasons}(a) illustrates the reasons for all orderings (both basic and transitive) in this plan, while Figure \ref{fig:ordering-reasons}(b) displays the reasons for only basic orderings.  
 \begin{verbatim}
1 (board p1 n2 e1)
2 (move_up e1 n2 n3)
3 (leave p1 n3 e1) 
4 (move_down e1 n3 n2)
5 (move_down e1 n2 n1)
6 (board p2 n1 e1)
7 (move_up e1 n1 n2)
8 (leave p2 n2 e1)
\end{verbatim}

Traditional plan deordering/reordering techniques cannot remove any ordering from this plan. The following section will demonstrate how block deordering eliminates orderings in this plan by forming blocks of operators. In our methodology, we solely focus on eliminating basic orderings in a plan, as removing transitive ordering does not enhance plan flexibility. In the rest of the paper, we will only show basic orderings in similar examples for clearer and simpler illustration.

\begin{figure}[!tbp]
    \centering
    \begin{subfigure}[t]{.72\textwidth}
    \centering
  \footnotesize
            \begin{tikzpicture}[node distance = 1cm]
            \node (init)[process] {\verb|INIT|};
            \node (op1)[process, below of = init, xshift =1.5cm, yshift=-.2cm]{\verb|board p1 n2 e1|};
            \node (op2)[process, xshift =-2.5cm, below of = op1]{\verb|move_up e1 n2 n3|};
            \node (op3)[process, below of = op2, xshift=3.5cm]{\verb|leave p1 n3 e1|};
            \node (op4)[process, below of = op3, xshift=-4.5cm]{\verb|move_down e1 n3 n2|};
             \node (op5)[process, below of = op4, xshift=-.5cm, yshift=-.2cm]{\verb|move_down e1 n2 n1|};
             \node (op6)[process, below of = op5, xshift=-2cm, yshift=-.2cm]{\verb|board p2 n1 e1|};
             \node (op7)[process, below of = op6, xshift = 5cm]{\verb|move_up e1 n1 n2|};
            \node (op8)[process, below of = op7]{\verb|leave p2 n2 e1|};
            \node (goal)[process, below of= op8] {\verb|GOAL|};
            
            \draw [arrow] (init) -> node[anchor=south west]{PC(\atom{$v_{e1}$}{n2})}node[anchor= west]{PC(\atom{$v_{p1}$}{n2})}(op1);
            \draw [arrow] (init) -> node[anchor= east]{PC(\atom{$v_{e1}$}{n2})}(op2);
            \draw [arrow] (op1.west) -> node[anchor=north west, xshift=-.3cm]{CD(\atom{$v_{e1}$}{n2})}(op2);
            \draw [arrow] (op1)  -> node[anchor=west]{PC(\atom{$v_{p1}$}{e1})}(op3);
            \draw [arrow] (op2.east) -> node[anchor=east, yshift=-.25cm, xshift=.5cm]{PC(\atom{$v_{e1}$}{n3})}(op3);
            \draw [arrow] (op2) -> node[anchor=east]{PC(\atom{$v_{e1}$}{n2})} node[anchor=south east]{DP(\atom{$v_{e1}$}{n1})}(op4);
            \draw [arrow] (op2.west) -- ++(-2.1cm, 0)|- node[anchor=west, xshift=-.1cm, yshift=1.8cm]{CD(\atom{$v_{e1}$}{n2})}(op5);
            \draw [arrow] (op2) -> node[anchor=west, yshift=-.7cm, xshift=.1cm]{DP(\atom{$v_{e1}$}{n3})}(op7);

    \draw [arrow] (init.west) -| node[anchor=west, yshift=-1cm]{PC(\atom{$v_{p2}$}{n1})}(op6);
            \draw [arrow] (op3) -> node[anchor=north west, xshift=-.4cm]{CD(\atom{$v_{e1}$}{n3})}(op4);
              \draw [arrow] (op3) |- node[anchor=east, yshift=5cm, xshift=.1cm]{PC(\atom{$v_{p1}$}{n3})}(goal);
           \draw [arrow] (op4) -> node[anchor= west]{PC(\atom{$v_{e1}$}{n2})}(op5);
           \draw [arrow] (op5) -> node[anchor=north west, xshift=-.3cm]{PC(\atom{$v_{e1}$}{n1})}(op6);
            \draw [arrow] (op5) -> node[anchor= east]{PC(\atom{$v_{e1}$}{n1})}(op7);
            \draw [arrow] (op6) ->node[anchor= east, yshift=-.2cm]{CD(\atom{$v_{e1}$}{n1})}(op7);
            \draw [arrow] (op6) |-node[anchor= west, yshift=.8cm]{PC(\atom{$v_{p2}$}{e1})}(op8);
           \draw [arrow] (op7) ->node[anchor= west]{PC(\atom{$v_{e1}$}{n2})}(op8);
            \draw [arrow] (op8) ->node[anchor= east]{PC(\atom{$v_{p2}$}{n2})}(goal); 
            \end{tikzpicture}
        \caption{}
    \end{subfigure}
    \begin{subfigure}[t]{.25\textwidth}
    \centering
    \footnotesize
     \begin{tikzpicture}[node distance = 1cm]
            \node (init)[process] {\verb|INIT|};
            \node (op1)[process, below of = init, yshift=-.2cm]{\verb|board p1 n2 e1|};
            \node (op2)[process, below of = op1]{\verb|move_up e1 n2 n3|};
            \node (op3)[process, below of = op2]{\verb|leave p1 n3 e1|};
            \node (op4)[process, below of = op3]{\verb|move_down e1 n3 n2|};
             \node (op5)[process, below of = op4]{\verb|move_down e1 n2 n1|};
             \node (op6)[process, below of = op5]{\verb|board p2 n1 e1|};
             \node (op7)[process, below of = op6]{\verb|move_up e1 n1 n2|};
            \node (op8)[process, below of = op7]{\verb|leave p2 n2 e1|};
            \node (goal)[process, below of= op8] {\verb|GOAL|};
            \draw [arrow] (init) -> node[anchor=south east]{PC(\atom{$v_{e1}$}{n2})}node[anchor= east]{PC(\atom{$v_{p1}$}{n2})}(op1);
            \draw [arrow] (op1) -> node[anchor=east]{CD(\atom{$v_{e1}$}{n2})}(op2);
            \draw [arrow] (op2) -> node[anchor=east]{PC(\atom{$v_{e1}$}{n3})}(op3);
            \draw [arrow] (op3) -> node[anchor=east]{CD(\atom{$v_{e1}$}{n3})}(op4);
           \draw [arrow] (op4) -> node[anchor= east]{PC(\atom{$v_{e1}$}{n2})}(op5);
           \draw [arrow] (op5) -> node[anchor= east]{PC(\atom{$v_{e1}$}{n1})}(op6);
            \draw [arrow] (op6) ->node[anchor= east]{CD(\atom{$v_{e1}$}{n1})}(op7);
           \draw [arrow] (op7) ->node[anchor= east]{PC(\atom{$v_{e1}$}{n2})}(op8);
            \draw [arrow] (op8) ->node[anchor= east]{PC(\atom{$v_{p2}$}{n2})}(goal);
        \end{tikzpicture}
        \caption{}
    \end{subfigure}
    
    \caption{A plan for the elevator problem instance, presented in Example \ref{example:sas}, annotated with ordering reasons according to Definition \ref{def:pc-cd-dp}: (a) for all orderings, including transitive orderings, and (b) for only basic orderings.}
    \label{fig:ordering-reasons}
\end{figure}
\end{example}


\subsection{Plan Deordering and Reordering}
Partial-order planning embodies the least commitment strategy, which aims to find flexible plans that allow delaying decisions during plan execution \cite{weld_1994}. Two essential concepts for achieving this flexibility are \emph{plan deordering} and \emph{reordering}. Following Bäckström \cite{backstrom1998}, Definition \ref{def:deordering} formally presents the notions of deordering and reordering of a POP, assuming that the POP is transitively closed.
 \begin{definition} \label{def:deordering}
     Let $P=\langle \mathcal{O}, \prec \rangle$ and $Q=\langle \mathcal{O}, \prec' \rangle$ be two partial-order plans for a planning task $\Pi$, then:
     \begin{itemize}
         \item $Q$ is a \textbf{deordering} of $P$  w.r.t. $\Pi$ iff $P$ and $Q$ are both valid POPs and $\prec' \subseteq \prec$.
          \item $Q$ is a \textbf{reordering} of $P$  w.r.t. $\Pi$ iff $P$ and $Q$ are both valid POPs.
         \item $Q$ is a \textbf{minimum deordering} of $P$  w.r.t. $\Pi$ iff 
         \begin{enumerate}
             \item $Q$ is a deordering of $P$, and
             \item there exists no POP $R=\langle \mathcal{O}, \prec ''\rangle$ s.t. $R$ is a deordering of $P$ and $|\prec''| < |\prec'|$.
         \end{enumerate}
     \end{itemize}
     \begin{itemize}
         \item $Q$ is a \textbf{minimum reordering} of $P$  w.r.t. $\Pi$ iff 
         \begin{enumerate}
             \item $Q$ is a reordering of $P$, and
             \item there exists no POP $R=\langle \mathcal{O}, \prec ''\rangle$ s.t. $R$ is a reordering of $P$ and $|\prec''| < |\prec'|$.
         \end{enumerate}
     \end{itemize}
 \end{definition}

Though \acrshort*{maxsat}-based reorderings \cite{maxsat,Waters_Nebel_Padgham_Sardina_2018,maxsat_reinst} can generate minimum reordering of a plan,  these techniques have high computational cost, which results in lower coverage. In contrast, deordering strategies, such as \acrfull*{eog} and block deordering, have trivial computational times. Thus, we use \acrshort*{eog} and block deordering to effectively minimize orderings in a plan. 

\subsubsection*{Explanation-based Order Generalization}
Explanation-based ordering generalization (EOG) \cite{kk,Veloso2002}  constructs a validation structure of a plan by establishing a causal link for every precondition of its operators. Then,  it adds \textit{promotion} or \textit{demotion} orderings to resolve any threats to the causal links. 

\begin{algorithm}[!tbp]
    \caption{EOG}
    \label{alg:kk_algorithm}
        \begin{algorithmic}[1]
            \State \textbf{Input:} a valid sequential plan $\pi =\langle o_1, \dots, o_n \rangle$
            \State \textbf{Output:} a valid partial-order plan
            \For {$1 < i \leq n$}   
            \Comment{Constructing validation structure}
                \For{$\langle v, d \rangle \in cons(o_i)$}
                    \State find min $k < i$ s.t.,
                    
                        \quad\quad1. $\langle v, d \rangle\in prod(o_k)$
                        
                        \quad\quad2. there is no $j$ s.t. $k <j <i$ and $\langle v, d \rangle\in  del(o_j)$.
             
                    \State add $o_k \xrightarrow{\langle v, d \rangle} o_i$ to $\prec$
                \EndFor 
            \EndFor
            \ForAll{$o_i,o_j \in \pi$ s.t. $ i < j $ } 
            \Comment{Resolving threats}
                    \State add $\langle o_i \prec o_j\rangle$ to $\prec$ if there exists an operator $o_k$, for which 
                    \Statex \quad\quad one of the following conditions is true,
                 
                    \quad$1.$ $o_k \xrightarrow{\langle v, d \rangle} o_i$ to $\prec$ and $o_j$ deletes the fact $\langle v, d \rangle$
                    
                    \quad$2.$ $o_j \xrightarrow{\langle v, d \rangle} o_k$ to $\prec$ and $o_i$ deletes the fact $\langle v, d \rangle$
            \EndFor
    \end{algorithmic}
\end{algorithm}

Let $\pi$ be a sequential plan for a planning task $\Pi= \langle \mathcal{V}, O, s_0, s_*\rangle$.
EOG, presented in Algorithm \ref{alg:kk_algorithm}, includes two extra operators  $o_I$ and $o_G$ in $\pi$ to replicate the initial state and goal conditions such that $pre(o_I)=\emptyset$, $\mathit{eff}(o_I) = s_0$,  $pre(o_G) = s_*$, $\mathit{eff}(o_G)=\emptyset$,  $o_I \prec o_G$ and for all operators $o \in (\pi\setminus \{o_I, o_G\})$, $o_I \prec o \prec o_G$. Then, it establishes the causal links (lines 3-8) and resolves threats (lines 9-11) using promotions and demotions.  This algorithm binds the earliest producers to causal links to reduce the chance of unnecessary transitive orderings.



\subsubsection*{Block Deordering}

Block deordering \cite{siddiqui_patrik_2012} removes ordering constraints in a POP by grouping coherent operators into blocks, thus converting the POP into a block decomposed partial-order (BDPO) plan. Following Noor and Siddiqui \cite{noor2024improving}, we use block semantics under FDR encoding.

A block encapsulates a set of operators, and a POP incorporating blocks is called a block decomposed partial-order (BDPO) plan. Operators of two disjoint blocks in a BDPO plan cannot interleave with each other, thus allowing the execution of the unordered blocks in any order. Blocks can also be nested, i.e., a block can contain one or more blocks, but they are not allowed to overlap.
\begin{definition} \label{def:block1}
A \textbf{block decomposed partial-order plan} is a 3-tuple  $\pi_{bdp} =\langle \mathcal{O}, \mathcal{B}, \prec\rangle$ where $\mathcal{O}$ is a set of operators, $\mathcal{B}$ is a set of blocks,  and $\prec$ is a set of ordering constraints over $\mathcal{O}$. Let $b \in \mathcal{B}$ be a block comprising a set of operators such that for any two operators $o, o' \in b$, where $o \prec o'$, there exists no other operator  $o'' \notin b$ with $o \prec o'' \prec o'$.
If $b_i, b_j \in \mathcal{B}$ are two blocks, then only one of these three relations, $b_i \subset b_j$, $b_j \subset b_i$, $b_i \cap b_j = \emptyset$ can be true. 
\label{def: bdpo}
\end{definition}

A block can be described by its precondition and its effects. A block $b$ has a fact \atom{v}{d} as its precondition if an operator $o\in b$ consumes \atom{v}{d}, and no other operator in $b$ provides \atom{v}{d} to the operator $o$. On the other hand, the block $b$ has the fact \atom{v}{d} as an effect if an operator $o\in b$ produces \atom{v}{d} and no other operator in $b$ that follows $o$ changes the value of $v$.

\begin{definition}\label{def:block}
    Let $\pi_{bdp} =\langle \mathcal{O}, \mathcal{B}, \prec\rangle$ be a BDPO plan, where $b \in \mathcal{B}$ be a block. The \textbf{block semantics} are defined as,
  \begin{itemize}
        \item A fact  $\langle v, d \rangle \in pre(b)$ iff there is an operator $o \in b$ with $\langle v, d \rangle \in pre(o)$, and $b$ has no other operator $o'$ such that there exists a causal link $o'\xrightarrow{\langle v, d \rangle } o$.
        \item A fact $\langle v, d \rangle \in \mathit{eff}(b)$ iff there exists an operator $o \in b$ with $\langle v, d \rangle  \in \mathit{eff}(o)$, and   no operator $o'\in b$ has an effect $\langle v, d' \rangle$ where $o \prec o'$ and $d' \in (\mathcal{D}_v\setminus \{d\})$.
    \end{itemize}
    \label{def: block-semantics}
\end{definition}
According to the Definition  \ref{def:block}, a block can have multiple facts over one variable as its effects. Assume, a block $b$ has two unordered operators $o$ and $o'$, where  $ \langle v, d \rangle \in \mathit{eff}(o)$ and $\langle v, d'\rangle \in \mathit{eff}(o')$ such that $d\ne d'$. In this situation, the block $b$ has both \atom{v}{d} and $\langle v, d'\rangle $ as its effects. This is to specify the facts that a block produces or deletes (Definition \ref{def1}).  

\begin{definition}
    \label{def1}
    The set of facts that are consumed, produced, and deleted by a block $b$ are denoted as \textbf{cons(b)}, \textbf{prod(b)}, and \textbf{del(b)}, respectively.
        \begin{itemize}
            \item  A fact $\langle v,d\rangle \in cons(b)$, iff $\langle v,d \rangle \in pre(b)$.
            \item  A fact $\langle v,d\rangle \in prod(b)$, iff 
            \begin{enumerate}
                \item $\langle v,d\rangle \notin cons(b)$,
                \item $\langle v,d \rangle \in \mathit{eff}(b)$, and
                \item $\langle v, d' \rangle \notin \mathit{eff}(b)$  where $d' \in (\mathcal{D}_v\setminus \{d\})$.
            \end{enumerate}
            \item A fact $\langle v,d\rangle \in del(b)$, iff
            \begin{enumerate}
                \item  Either $v \notin \mathit{vars}(\mathit{cons}(b))$ or  $cons(b)(v) = d$, and 
                \item $\langle v, d' \rangle \in \mathit{eff}(b)$, where $d' \in (\mathcal{D}_v\setminus \{d\})$. 
            \end{enumerate}
        \end{itemize}
\end{definition}
A block $b$ is not considered to produce a fact $\langle v, d \rangle$ when it also consumes \atom{v}{d}. The intuition behind this is to perceive a block as an operator, and to insist that \atom{v}{d} persists before and after the block's execution. 

The notation $b_i \prec b_j$, where $b_i$ and $b_j$ are two blocks in a BDPO plan, indicates that there exist two operators $o$ and $o'$ such that $o\in b_i$, $o'\in b_j$, and $o\prec o'$. We can also use the $PC, CD$, and $DP$ labels (Definition \ref{def:pc-cd-dp}) to annotate orderings in a BDPO plam. The following definition presents the concepts of \emph{candidate producer} and \emph{earliest candidate producers} for a block's precondition. These terms will be used in the forthcoming algorithms.

\begin{definition}\label{def:candidate_producer}
   Let $\pi_{bdp} =\langle \mathcal{O}, \mathcal{B}, \prec\rangle$ be a BDPO plan, and  $b_i, b_j \in \mathcal{B}$ be two blocks, where $\langle v, d \rangle \in cons(b_j)$,
   \begin{itemize}
       \item Block $b_i$ is a \textbf{candidate producer} of a fact $\langle v, d \rangle$ for $b_j$ if
       \begin{enumerate}
           \item  $\langle v, d \rangle \in \mathit{prod}(b_i)$,
           \item $b_i \prec b_j$, and there exists no block  $b_k \in \mathcal{B}$ with $ \langle v, d \rangle\in del(b_k)$, where  $b_j \nprec b_k \nprec b_i$. 
       \end{enumerate}
           
     \item Block $b_i$ is the \textbf{earliest candidate producer} of a fact $\langle v, d \rangle$ for $b_j$ if there is no candidate producer $b_k$ of $\langle v, d \rangle$ for $b_j$ such that $b_k \prec b_i$.
    \end{itemize}
\end{definition}


Block deordering takes a sequential plan as input, and produces a valid BDPO plan. It begins by transforming the sequential plan into a POP $\pi=\langle \mathcal{O}, \prec \rangle$ using EOG. Then it constructs an initial BDPO plan $\pi_{bdp} = (\mathcal{O}, \mathcal{B}, \prec)$ simply by adding a block $b=\{o\}$ to $\mathcal{B}$ for each operator $o\in\mathcal{O}$. Also, for every $ o_i \prec o_j$ in $\prec$,  it adds an ordering $b_i \prec b_j$ where $o_i\in b_i, o_j\in b_j$ and $b_i, b_j \in \mathcal{B}$.  Then, block deordering applies the following rule \cite{siddiqui_patrik_2012} to eliminate orderings in $\pi_{bdp}$. We use the terms \emph{primitive} and \emph{compound} blocks to specify blocks with single and multiple operators, respectively.  When describing rules and algorithms, the term \emph{block} is used to refer to both primitive and compound blocks generally.

\begin{rle}
    \label{rule_1}
    Let  $\pi_{bdp} =\langle \mathcal{O}, B \prec \rangle$ be a valid BDPO plan, and  $b_i \prec b_j$ be a basic ordering,
    \begin{enumerate}[label=\roman*.]
        \item 
        \label{rule_1a} Let $PC(\langle v, d\rangle) \in Re(b_i\prec b_j)$ be a ordering reason, and $b$ be a block, where $b_i \in b, b_j \notin b$ and $\forall b' \in \{b\setminus b_i\}, b_i \nprec b'$. $PC(\langle v, d\rangle)$ can be removed from $Re(b_i\prec b_j)$ if $\langle v, d\rangle \in pre(b)$ and $\exists b_p \notin b$ such that $b_p$ can establish causal links $b_p \xrightarrow{\langle v, d\rangle} b_j $ and $b_p \xrightarrow{\langle v, d\rangle} b$.
        
        \item \label{rule_1b}  Let $CD(\langle v, d\rangle) \in Re(b_i\prec b_j)$ be a ordering reason, and $b$ be a block, where $b_i \in b, b_j \notin b$ and $b \cap b_j = \emptyset$. Then $CD(\langle v, d\rangle)$ can be removed from $Re(b_i\prec b_j)$ if $b$ does not consume $\langle v, d\rangle$.
        
        \item \label{rule_1c} Let $CD(\langle v, d\rangle) \in Re(b_i\prec b_j)$ be a ordering reason, and $b$ be a block, where $b_i \notin b, b_j\in b$ and $b_i \cap b = \emptyset$. The $CD(\langle v, d\rangle)$ can be removed from $Re(b_i\prec b_j)$  if $b$ does not delete $\langle v, d\rangle$.
        \item\label{rule_1d} Let $DP(\langle v, d\rangle) \in Re(b_i\prec b_j)$ be a ordering reason, and $b$ be a block, where, $b_j \in b$, but $b_i \notin b$. Then $DP(\langle v, d\rangle)$ can be removed from $Re(b_i\prec b_j)$ if $b$ includes all blocks $b'$ such that $b_j \xrightarrow{\langle v, d\rangle} b'$. 
    \end{enumerate}
\end{rle}

To remove a $PC(\langle v, d\rangle)$ reason  from $Re(b_i\prec b_j)$, Rule 1(i) searches for a block $b_c$ such that $b_c \prec b_i$ and $\langle v, d \rangle \in cons(b_c)$. If $b_c$ is found, it forms a block $b$ encapsulating $b_i$, $b_c$, and all the blocks ordered between $b_c$ and $b_i$. Since $b_c$ consumes  $\langle v, d \rangle$, there must be a block $b_p$ such that $b_p \xrightarrow{\langle v, d \rangle} b_c$. Therefore, Rule 1(i) establishes $b_p \xrightarrow{\langle v, d \rangle} b$ and $b_p \xrightarrow{\langle v, d \rangle} b_j$, allowing  $PC(\langle v, d\rangle)\in Re(b_i\prec b_j)$ reason to be removed. To eliminate $CD(\langle v, d\rangle) \in Re(b_i\prec b_j)$,  Rule 1(ii-iii) seeks a block $b_p$ with $\langle v, d\rangle \in prod(b_p)$.  If $b_p$ precedes $b_i$ and then Rule 1(ii) creates a new block $b$ with blocks $b_p$, $b_i$ and every block $b'$ such that $b_p \prec b'\prec b_i$. On the other hand, if $b_p$ follows $b_j$, Rule 1(iii) forms the new block $b$ by encompassing $b_j$, $b_p$, and every block $b'$ s.t. $b_j \prec b' \prec b_p$. For removing $DP(\langle v, d\rangle) \in Re(b_i\prec b_j)$,  Rule 1(iv) forms a new block $b$  that includes $b_j$, every block $b'$ such that $b_j \xrightarrow{\langle v,d \rangle} b'$, and each block $b''$ with $b_i\prec b''\prec b'$. The new block $b$ functions as a barrier against the corresponding deleter. These rules have been further illustrated in the work by Siddiqui and Patrik \cite{siddiqui_patrik_2012}.

 Block deordering attempts to eliminate each basic ordering from the top of the initial BDPO plan. An ordering $b_i \prec b_j$ is removed if all of its ordering reasons can be eliminated by applying Rule 1. The ordering $b_i \prec b_j$ remains if any of its reasons can not be eliminated. If an attempt to eliminate an ordering is unsuccessful, the algorithm proceeds to the next ordering. When the algorithm successfully removes an ordering, it returns the newly generated BDPO plan and restarts the deordering process from the top of this latest BDPO plan. This iterative process continues until no further deordering is possible with the most recent BDPO plan. 

 \begin{example}\label{example:bd}
  \begin{figure}[!tbp]
    \centering
    \begin{subfigure}[t]{.45\textwidth}
    \centering
    \footnotesize
     \begin{tikzpicture}[node distance = 1cm]
            \node (init)[process] {\verb|INIT|};
            \node (op1)[process, below of = init, xshift=.2cm]{\verb|board p1 n2 e1|};
            \node (op2)[process, below of = op1]{\verb|move_up e1 n2 n3|};
            \node (op3)[process, below of = op2]{\verb|leave p1 n3 e1|};
            \node (op4)[process, below of = op3]{\verb|move_down e1 n3 n2|};
             \node (op5)[process, below of = op4]{\verb|move_down e1 n2 n1|};
             \node (op6)[process, below of = op5]{\verb|board p2 n1 e1|};
             \node (op7)[process, below of = op6]{\verb|move_up e1 n1 n2|};
            \node (op8)[process, below of = op7]{\verb|leave p2 n2 e1|};
            \node (goal)[process, below of= op8, xshift=-.15cm] {\verb|GOAL|};

    \begin{pgfonlayer}{background}
            \node (b1) [fit=(op1)] [block, label={[xshift=5.5mm]left:{\small{$b_1$}}}] {};
            \node (b2) [fit=(op2)] [block, label={[xshift=5.5mm]left:{\small{$b_2$}}}] {};
            \node (b3) [fit=(op3)] [block, label={[xshift=5.5mm]left:{\small{$b_3$}}}] {};
            \node (b4) [fit=(op4)] [block, label={[xshift=5.5mm]left:{\small{$b_4$}}}] {};
            \node (b5) [fit=(op5)] [block, label={[xshift=5.5mm]left:{\small{$b_5$}}}] {};
            \node (b6) [fit=(op6)] [block, label={[xshift=5.5mm]left:{\small{$b_6$}}}] {};
            \node (b7) [fit=(op7)] [block, label={[xshift=5.5mm]left:{\small{$b_7$}}}] {};
            \node (b8) [fit=(op8)] [block, label={[xshift=5.5mm]left:{\small{$b_8$}}}] {};        
        \end{pgfonlayer}

            \draw [arrow] (init) -> node[anchor=west]{PC(\atom{$v_{e1}$}{n2})}node[anchor= east]{PC(\atom{$v_{p1}$}{n2})}(b1);
            \draw [arrow] (b1) -> node[anchor=east]{CD(\atom{$v_{e1}$}{n2})}(b2);
            \draw [arrow] (b2) -> node[anchor=east]{PC(\atom{$v_{e1}$}{n3})}(b3);
            \draw [arrow] (b3) -> node[anchor=east]{CD(\atom{$v_{e1}$}{n3})}(b4);
           \draw [arrow] (b4) -> node[anchor= east]{PC(\atom{$v_{e1}$}{n2})}(b5);
           \draw [arrow] (b5) -> node[anchor= east]{PC(\atom{$v_{e1}$}{n1})}(b6);
            \draw [arrow] (b6) ->node[anchor= east]{CD(\atom{$v_{e1}$}{n1})}(b7);
           \draw [arrow] (b7) ->node[anchor= east]{PC(\atom{$v_{e1}$}{n2})}(b8);
            \draw [arrow] (b8) ->node[anchor= east]{PC(\atom{$v_{p2}$}{n2})}(goal);
        \end{tikzpicture}
        \caption{}
    \end{subfigure}
    \begin{subfigure}[t]{.45\textwidth}
    \centering
  \footnotesize
         \begin{tikzpicture}[node distance = 1cm]
            \node (init)[process] {\verb|INIT|};
            \node (op1)[process, below of = init, xshift=.2cm]{\verb|board p1 n2 e1|};
            \node (op2)[process, below of = op1, yshift=-.2cm, xshift=.1cm]{\verb|move_up e1 n2 n3|};
            \node (op3)[process, below of = op2]{\verb|leave p1 n3 e1|};
            \node (op4)[process, below of = op3]{\verb|move_down e1 n3 n2|};
             \node (op5)[process, below of = op4, yshift=-.2cm, xshift=-.1cm]{\verb|move_down e1 n2 n1|};
             \node (op6)[process, below of = op5]{\verb|board p2 n1 e1|};
             \node (op7)[process, below of = op6]{\verb|move_up e1 n1 n2|};
            \node (op8)[process, below of = op7]{\verb|leave p2 n2 e1|};
            \node (goal)[process, below of= op8, xshift=-.15cm] {\verb|GOAL|};
            
        \begin{pgfonlayer}{background}
            \node (b1) [fit=(op1)] [block, label={[xshift=5.5mm]left:{\small{$b_1$}}}] {};
            \node (b2) [fit=(op2)] [block, label={[xshift=5.5mm]left:{\small{$b_2$}}}] {};
            \node (b3) [fit=(op3)] [block, label={[xshift=5.5mm]left:{\small{$b_3$}}}] {};
            \node (b4) [fit=(op4)] [block, label={[xshift=5.5mm]left:{\small{$b_4$}}}] {};
            \node (b5) [fit=(op5)] [block, label={[xshift=5.5mm]left:{\small{$b_5$}}}] {};
            \node (b6) [fit=(op6)] [block, label={[xshift=5.5mm]left:{\small{$b_6$}}}] {};
            \node (b7) [fit=(op7)] [block, label={[xshift=5.5mm]left:{\small{$b_7$}}}] {};
            \node (b8) [fit=(op8)] [block, label={[xshift=5.5mm]left:{\small{$b_8$}}}] {};
             \node (fit1) [fit=(b2) (b3) (b4)] [block, fill=none, label={[xshift=5.5mm]left:{\small{$\mathbf{b_i}$}}}] {};
             
        \end{pgfonlayer}
        \draw [arrow] (init) -> node[anchor=south east, xshift=-.3cm]{PC(\atom{$v_{e1}$}{n2})}node[anchor= east]{PC(\atom{$v_{p1}$}{n2})}(b1);
             \draw [arrow, ultra thick] (init) -- ++(2.4cm,0) |- node[anchor= east, yshift=5.2cm]{\textbf{PC(\atom{$v_{e1}$}{n2})}}(b5.east);
            \draw [arrow, ultra thick] (b1) -> node[anchor=east]{\textbf{PC(\atom{$v_{p1}$}{e1})}}(fit1);
            \draw [arrow] (b2) -> node[anchor=east]{PC(\atom{$v_{e1}$}{n3})}(b3);
            \draw [arrow] (b3) -> node[anchor=east]{CD(\atom{$v_{e1}$}{n3})}(b4);
           
           \draw [arrow] (b5) -> node[anchor= east]{PC(\atom{$v_{e1}$}{n1})}(b6);
            \draw [arrow] (b6) ->node[anchor= east]{CD(\atom{$v_{e1}$}{n1})}(b7);
           \draw [arrow] (b7) ->node[anchor= east]{PC(\atom{$v_{e1}$}{n2})}(b8);
            \draw [arrow] (b8) ->node[anchor= west]{PC(\atom{$v_{p2}$}{n2})}(goal);

        \draw [arrow, ultra thick] (fit1) -> node[anchor= east]{\textbf{CD(\atom{$v_{e1}$}{n2})}}(b5);

        \end{tikzpicture}
        \caption{}
    \end{subfigure}
    \begin{subfigure}[t]{.8\textwidth}
    \centering
    \footnotesize
     \begin{tikzpicture}[node distance = 1cm]
            \node (init)[process] {\verb|INIT|};
            \node (op1)[process, below of = init, yshift=-.2cm, xshift=-2.5cm]{\verb|board p1 n2 e1|};
            \node (op2)[process, below of = op1, yshift=-.2cm, xshift=.1cm]{\verb|move_up e1 n2 n3|};
            \node (op3)[process, below of = op2]{\verb|leave p1 n3 e1|};
            \node (op4)[process, below of = op3]{\verb|move_down e1 n3 n2|};
             \node (op5)[process, below of = init, xshift=2.5cm, yshift=-.2cm]{\verb|move_down e1 n2 n1|};
             \node (op6)[process, below of = op5]{\verb|board p2 n1 e1|};
             \node (op7)[process, below of = op6]{\verb|move_up e1 n1 n2|};
            \node (op8)[process, below of = op7, yshift=-.15cm]{\verb|leave p2 n2 e1|};
            \node (goal)[process, below of= op8, xshift=-2.5cm, yshift=-.2cm] {\verb|GOAL|};

        \begin{pgfonlayer}{background}
            \node (b1) [fit=(op1)] [block, label={[xshift=5.5mm]left:{\small{$b_1$}}}] {};
            \node (b2) [fit=(op2)] [block, label={[xshift=5.5mm]left:{\small{$b_2$}}}] {};
            \node (b3) [fit=(op3)] [block, label={[xshift=5.5mm]left:{\small{$b_3$}}}] {};
            \node (b4) [fit=(op4)] [block, label={[xshift=5.5mm]left:{\small{$b_4$}}}] {};
            \node (b5) [fit=(op5)] [block, label={[xshift=5.5mm]left:{\small{$b_5$}}}] {};
            \node (b6) [fit=(op6)] [block, label={[xshift=5.5mm]left:{\small{$b_6$}}}] {};
            \node (b7) [fit=(op7)] [block, label={[xshift=5.5mm]left:{\small{$b_7$}}}] {};
            \node (b8) [fit=(op8)] [block, label={[xshift=5.5mm]left:{\small{$b_8$}}}] {};
             \node (fit1) [fit=(b2) (b3) (b4)] [block, fill=none, label={[xshift=5.5mm]left:{\small{$\mathbf{b_i}$}}}] {};
             \node (fit2) [fit=(b5) (b6) (b7)] [block, ultra thick, fill=none, label={[xshift=5.5mm]left:{\small{$\mathbf{b_j}$}}}]{};
        \end{pgfonlayer}
            
            \draw [arrow] (init) -> node[anchor=south east]{PC(\atom{$v_{e1}$}{n2})}node[anchor= east]{PC(\atom{$v_{p1}$}{n2})}(b1);
            \draw [arrow] (b1) -> node[anchor=east]{PC(\atom{$v_{p1}$}{e1})}(fit1);
            \draw [arrow] (b2) -> node[anchor=east]{PC(\atom{$v_{e1}$}{n3})}(b3);
            \draw [arrow] (b3) -> node[anchor=east]{CD(\atom{$v_{e1}$}{n3})}(b4);
           
           \draw [arrow] (b5) -> node[anchor= east]{PC(\atom{$v_{e1}$}{n1})}(b6);
            \draw [arrow] (b6) ->node[anchor= east]{CD(\atom{$v_{e1}$}{n1})}(b7);
           \draw [arrow] (fit2) ->node[anchor= east]{PC(\atom{$v_{p2}$}{e1})}(b8);
            \draw [arrow] (b8) ->node[anchor= west]{PC(\atom{$v_{p2}$}{n2})}(goal);
        
        \draw [arrow] (fit1) -> node[anchor= east]{PC(\atom{$v_{p1}$}{n3})}(goal);
        \draw [arrow] (init) -> node[anchor= south west]{PC(\atom{$v_{e1}$}{n2})}node[anchor= west]{PC(\atom{$v_{p2}$}{n2})}(fit2);
        \end{tikzpicture}
        \caption{}
    \end{subfigure} 
    \caption{Removing the ordering $b4\prec b5$ from the BDPO plan, presented in (a), using Rule \ref{rule_1}. (b) Rule 1\ref{rule_1a} forms block $b_i$  to remove the reason $PC(\langle v_{e1}, n2\rangle)\in Re(b_4\prec b_5)$.
   (c) Rule 1\ref{rule_1c} then forms another block $b_j$ to remove reason $CD(\langle v_{e1}, n2\rangle)\in Re(b_i\prec b_5)$, yielding a BDPO plan with no ordering between $b_i$ and $b_j$, where $b_4\in b_i$ and $b_5\in b_j$.}
    \label{fig:bd}
\end{figure}
 Let us recall the plan from Example \ref{example:pop}, presented in Figure \ref{fig:ordering-reasons}(b).
 Figure \ref{fig:bd}(a) interprets that plan as an initial BDPO plan by adding a primary block for each operator. Blocks $b_4$ and $b_5$ contain operators \verb|move_down e1 n3 n2| and \verb|move_down e1 n2 n1|, respectively. Block $b_4$ produces the fact $\langle v_{e1}, n2\rangle$, which is consumed by block $b_5$, forming an ordering reason $PC(\langle v_{e1}, n2\rangle)\in Re(b_4\prec b_5)$. To remove this reason, Rule 1\ref{rule_1a} finds a block $b_2$ that consumes the fact $\langle v_{e1}, n2\rangle$, where $b_2\prec b_4$. Then it creates a block $b_i$ by encapsulating blocks $b_2, b_3$ and  $b_4$, resulting in a BDPO plan presented in Figure \ref{fig:bd}(b). Notably, both this latest BDPO plan and block $b_i$ is temporarily formed, because block $b_i$, containing $b_4$, still has an ordering with $b_5$ due to the reason $CD(\langle v_{e1}, n2\rangle)$. If every ordering reason between blocks $b_4$ and $b_5$ can not be removed, then this process rolls back to the original plan, shown in \ref{fig:bd}(a). Hence, Rule 1\ref{rule_1c} is then applied to remove the reason $CD(\langle v_{e1}, n2\rangle)\in Re(b_i\prec b_5)$. It finds a block $b_7$, producing the fact $\langle v_{e1}, n2\rangle$, and creates a new block $b_j$ over blocks $b_5$ to $b_7$, yielding a BDPO plan (presented in Figure \ref{fig:bd}(c)) having no ordering between $b_i$ and $b_j$, where $b_4\in b_i$ and $b_5\in b_j$.
 \end{example}

BDPO plans can be further post-processed by block-substitution \cite{noor2024improving}, which substitutes a block in a BDPO plan with a subplan w.r.t. the planning task. This work employs both block deordering and block substitution in the proposed methodology, presented in Section \ref{sec:cibs}.

\subsection{Block Substitution}\label{sec:block-substitution}
Block-substitution \cite{noor2024improving} allows substituting a block in a valid BDPO plan while maintaining the plan's validity. The block being replaced is referred to as the original block, and the block that replaces it is called the substituting block. The block-substitution process creates causal links for the preconditions of the substituting block and reestablishes all causal links that were previously supported by the original block. Additionally, this process must resolve any potential threats introduced by the substitution to ensure the plan remains valid.

\begin{definition}
    Let $\pi_{bdp}=\langle \mathcal{O}, \mathcal{B}, \prec \rangle$ be a valid BDPO plan with respect to a planning task $\Pi =\langle \mathcal{V}, O, s_0, s_* \rangle$, and $b\in \mathcal{B}$ be a block.  Let $\hat{b}=\langle \hat{O},\hat{\prec}\rangle$ be a partial-order subplan w.r.t. $\Pi$. A \textbf{block-substitution} of $b$ with $\hat{b}$ yields a  BDPO plan $\pi'_{bdp}= \langle \mathcal{O'}, \mathcal{B'}, \prec'\rangle$, where $b\notin \mathcal{B}'$, and $\hat{b}\in \mathcal{B}'$. A block-substitution is valid when $\pi'_{bdp}$ is valid.
\end{definition}

\begin{figure}[!bt]

\begin{subfigure}{.48\columnwidth}
     \centering
        \begin{tikzpicture}[node distance= 1.2cm]
            \node (init)[process]{INIT};
            \node (r)[block2, below of= init,]{$b_r$};
            \node (i)[block2, below of= r]{$b_i$};
            \node (s)[block2, below of= i,  xshift=1cm, yshift=-.5cm]{$b_s$};
            \node (x)[block2, below of= i, xshift=-1cm]{$b_x$};
            \node (t)[block2, below of= x]{$b_t$};
            \node (goal)[process, below of= t,xshift=1cm ]{GOAL};
           
            \draw [arrow, dotted](init) -> (r);
            \draw [arrow](r) -> node[right]{$PC(\langle v_1, d_1\rangle)$} (i);
            \draw [arrow](i) ->node[left]{$PC(\langle v_2, d_2\rangle)$}  (x);
            \draw [arrow](x) -> node[left]{$PC(\langle v_3, d_3\rangle)$} (t);
            \draw [ arrow, dotted ](t) -> (goal);
            \draw [ arrow, dotted](s) ->  (goal);
            \draw [arrow](i) -> node[right]{$CD(\langle v_1, d_1\rangle)$} (s);
        \end{tikzpicture}
        \caption{}
        \label{fig:substitution-example-1a}
\end{subfigure}
\begin{subfigure}{.48\columnwidth}
     \centering
        \begin{tikzpicture}[node distance= 1.2cm]
            \node (init)[process]{INIT};
            \node (r)[block2, below of= init]{$b_r$};
            \node (i)[block2, below of= r]{$b_i$};
            \node (s)[block2, below of= i,  xshift=1.2cm, yshift =-.5cm]{$b_s$};
            \node (x)[block2, below of= i, xshift=-1.4cm]{$\hat{b}_x$};
            \node (t)[block2, below of= x]{$b_t$};
            \node (goal)[process, below of= t,xshift=1.2cm ]{GOAL};
           
            \draw [arrow, dotted](init) -> (r);
            \draw [arrow](r) -> node[right]{$PC(\langle v_1, d_1\rangle)$} (i);
            \draw [arrow](r) -> node[left]{$PC(\langle v_1, d_1\rangle)$}  (x);
            \draw [arrow](x) -> node[left]{$PC(\langle v_3, d_3\rangle)$} (t);
            \draw [ arrow, dotted ](t) -> (goal);
            \draw [ arrow, dotted](s) ->  (goal);
            \draw [arrow](i) -> node[right]{$CD(\langle v_1, d_1\rangle)$} (s);
             \draw [arrow](x) -> node[left, yshift =-.3cm, xshift=1cm]{$(CD\langle v_1, d_1\rangle)$} (s);

        \end{tikzpicture}
        \caption{}
        \label{fig:substitution-example-1b}
\end{subfigure}
\caption{Substituting a block $b_x$ in (a) a valid BDPO plan $\pi_{bdp}=\langle \mathcal{O},\mathcal{B}, \prec\rangle$  with a block $\hat{b}_x \notin \mathcal{B}$, where $\langle v_1, d_1\rangle \in cons(\hat{b}_x)$ and $\langle v_3, d_3\rangle \in prod(\hat{b}_x)$. (b) This substitution adds two causal links $b_r \xrightarrow{\langle v_1, d_1 \rangle} \hat{b}_x$ and $\hat{b}_x \xrightarrow{\langle v_3, d_3 \rangle} b_t$, and an ordering reason $CD(\langle v_1, d_1 \rangle)$ to $Re(\hat{b}_x \prec b_s)$ for resolving threat, producing a valid BDPO plan where blocks $b_i$ and $\hat{b}_x$ are unordered.  The dotted lines represent ordering (basic or transitive) between two blocks. }
\label{fig:substitution-example-1}
\end{figure}

\begin{example}
Let us consider the BDPO plan $\pi_{bdp}=\langle \mathcal{O},\mathcal{B}, \prec\rangle$ in Figure \ref{fig:substitution-example-1}(a). In this plan, the precondition of block $b_x$ is supported by the causal link $b_i \xrightarrow{\langle v_2, d_2 \rangle} b_x$, and $b_x$ provides $\langle v_3, d_3 \rangle$ to block $b_t$. Block $b_s$ deletes $\langle v_1, d_1 \rangle$ and threatens $b_r \xrightarrow{\langle v_1, d_1\rangle} b_i$. The $CD(\langle v_1, d_1\rangle)$ ordering is added to $\in Re(b_i \prec b_s)$ to resolve this threat.  Let $\hat{b}_x\notin \mathcal{B}$ be a block, where $\langle v_1, d_1\rangle \in cons(\hat{b}_x)$ and $\langle v_3, d_3 \rangle \in prod(\hat{b}_x)$.  Substituting the block $b_x$ with $\hat{b}_x$ requires establishing causal links for the precondition of $\hat{b}_x$ and reestablishing the causal link $b_x\xrightarrow{\langle v_3, d_3\rangle} b_t$. Therefore, causal links $b_r \xrightarrow{\langle v_1, d_1\rangle} \hat{b}_x$ and $\hat{b}_x \xrightarrow{\langle v_3, d_3\rangle} b_t$ are added to the resultant BDPO plan after substituting $b_x$ with $\hat{b}_x$ (shown in Figure \ref{fig:substitution-example-1b}). However, $b_s$ becomes a threat to  $b_r \xrightarrow{\langle v_1, d_1\rangle} \hat{b}_x$ as  $b_s$ deletes $\langle v_1, d_1\rangle$. An ordering reason $CD(\langle v_1, d_1\rangle)$ is added to $Re(\hat{b}_x \prec b_s)$ to resolve this threat. Since each block's preconditions are now supported by a causal link, and no threats persist, replacing $b_x$ with $\hat{b}_x$ successfully produces a valid BDPO plan. Notably, the resultant plan contains no ordering between $b_r$ and $\hat{b}_x$.
\end{example}

Block-substitution allows sourcing the substituting block from within or outside the plan. We refer to the substitution as an internal block-substitution when the substituting block is from within the plan.  


Let  $b_t\in \mathcal{B}$ be a threat to a causal link $b_i \xrightarrow{\langle v, d \rangle} b_j$ in $\pi_{bdp}$, where $\langle v, d \rangle \in del(b_t)$ and $b_i, b_j \in \mathcal{B}$. The situations where $b_t$ pose a threat to a causal link are thoroughly examined in \cite{noor2024improving}. This threat can be resolved if any of the following \textbf{threat-resolving strategies} can be employed without introducing any cycle in $\pi_{bdp}$.
\begin{enumerate}[label=\arabic*.]
    \item \textit{Promotion}: adding an ordering $b_t \prec b_i$ to $\prec$.
    \item  \textit{Demotion}: adding an ordering $b_j \prec b_t$ to $\prec$.
    \item  \textit{Internal block-substitution}: substituting $b_j$ with $b_t$ or substituting $b_t$ with $b_j$.
\end{enumerate}

\begin{algorithm}[!t]
    \caption{Substitituting a block in a block decomposed partial-order (BDPO) plan}
    \label{alg:block_substitute}
    \textbf{Input}: a BDPO plan $\pi_{bdp}=\langle \mathcal{O},\mathcal{B}, \prec\rangle$, two block $b_x\in \mathcal{B}$ and $\hat{b}_x$.\\
    \textbf{Output:} a BDPO plan and a boolean value indicating whether the substitution is successful
    \begin{algorithmic}[1] 
    \Procedure{Substitute}{$\pi_{bdp}, b_x, \hat{b}_x$}
    \State $\hat{\pi}_{bdp}\equiv \langle \mathcal{O}', \mathcal{B}', \prec' \rangle\leftarrow \pi_{bdp}$
    
    \If{$\hat{b}_x\notin \mathcal{B}'$}
    \Comment{establishing causal links for $\hat{b}_x$'s precondition}
        \State $b_{new}\leftarrow \hat{b}_x$ 
        \State add $\hat{b}_x$ to $\mathcal{B}'$
        \ForAll{$\langle v, d \rangle\in \mathit{pre}(\hat{b}_x)$}
            \State find an earliest candidate producer $b$ of $\langle v, d \rangle$  for $\hat{b}_x$
            \If{$b$ is found}
             add $b \xrightarrow{\langle v, d \rangle} \hat{b}_x$ to $\prec'$
            \Else \ \Return $\pi_{bdp}$, \FALSE 
            \EndIf
        \EndFor
    \EndIf
     
        \ForAll{ $b\in \mathcal{B}'$ s.t. $b_x \xrightarrow{\langle v, d \rangle} b$} 
        \Comment{reestablishing causal links}
            \If{$\hat{b}_x$ produces $\langle v, d \rangle$}
            add $\hat{b}_x\xrightarrow{\langle v, d \rangle} b$ to $\prec'$
            \Else \ \Return $\pi_{bdp}$, \FALSE 
            \EndIf
        \EndFor
       \State delete $b_x$ from $\mathcal{B}'$
        
        \ForAll{threats where $b_k$ threatens $b_i\xrightarrow{\langle v, d\rangle} b_j$ s.t  $b_i, b_j, b_k\in \mathcal{B}'$}

        \If{$b_k\nprec b_j$} $\eta \leftarrow b_j\prec b_k$
            \Comment{demotion ordering}
            \Else\ $\eta \leftarrow b_k\prec b_i$ \Comment{promotion ordering}
        \EndIf
        \If{adding $\eta$ to $\prec'$ renders no cycle in $\hat{\pi}_{bdp}$}
          add $\eta$ to $\prec'$ 
        \Else 
        \Comment{try internal substitution by $\hat{b}_x$}
           \If{$b_k = b_{new}$}
            ($\hat{\pi}_{bdp}, success$)$\leftarrow$ \Call{Substitute}{$\hat{\pi}_{bdp}$, $b_j, b_k$}
            \ElsIf{$b_j = b_{new}$}
            \State ($\hat{\pi}_{bdp}, success$)$\leftarrow$ \Call{Substitute}{$\hat{\pi}_{bdp}$, $b_k, b_j$}
            \Else \ \Return $\pi_{bdp}$, \FALSE 
            \EndIf
             \If{$success$ is \FALSE}  
              \Return $\pi_{bdp}$, \FALSE 
             \EndIf
        \EndIf
        \EndFor
        \State \Return $\hat{\pi}_{bdp}$, \TRUE 
    \EndProcedure
   
    \end{algorithmic}
\end{algorithm}

Promotion or demotion orderings can invalidate a plan if it induces any cycle in that plan. In such scenarios, the threat can be resolved by an internal substitution, if possible. During block-substitution, if a threat cannot be resolved by any of these strategies, the substitution is not valid.

Algorithm \ref{alg:block_substitute} presents  the block-substitution procedure, named SUBSTITUTE. The inputs of the SUBSTITUTE procedure are  a valid BDPO plan $\pi_{bdp}=\langle \mathcal{O}, \mathcal{B}, \prec \rangle$, and two blocks $b_x$ and $\hat{b}_x$ s.t. $b_x\in \mathcal{B}$.  The block $\hat{b}_x$ can be sourced either from inside or outside  $\pi_{bdp}$. 

We can divide this procedure into three main parts. It first ensures that a causal link supports each precondition of $\hat{b}_x$.  When the block $\hat{b}_x$ is from the plan(i.e., $\hat{b}_x \in \mathcal{B}$), its preconditions are already supported by causal links. On the other hand, when $\hat{b}_x$ is from outside the plan, i.e.,  $\hat{b}_x \notin \mathcal{B}$, the procedure adds $\hat{b}_x$ to $\pi_{bdp}$. Then it forms a causal link with the earliest candidate producer (Definition \ref{def:candidate_producer}) for each precondition of  $\hat{b}_x$ (lines 3-12). If this process fails to establish a causal link for a precondition of $\hat{b}_x$, the substitution becomes unsuccessful. Afterward, the procedure reinstates the causal links, previously supported by the original block $b_x$, using the substituting block $\hat{b}_x$ as the new producer (lines 13-17). However, if $\hat{b}_x$ does not have any fact $\langle v,d \rangle\in prod(\hat{b}_x)$ such that $b_x \xrightarrow{\langle v,d \rangle} b$ for a block $ b\in \mathcal{B}$, the substitution is considered unsuccessful. After establishing all the required causal links, the block $b_x$ is removed from the plan. 

Next, the procedure identifies the threats introduced during this process and employs threat-resolving strategies to handle them (lines 19-32). If the procedure successfully resolves all threats, it returns the resultant BDPO plan. When we resolve threats using internal substitutions, we only consider substituting the conflicting block with $\hat{b}_x$, assuming that the conflicting block in the resultant BDPO plan becomes redundant only after substituting $b_x$ with $\hat{b}_x$.

\begin{theorem} \label{theorem:substitution}
(Correctness of Block-Substitution Algorithm) Given a valid BDPO plan $\pi_{bdp}=\langle \mathcal{O},\mathcal{B} \prec\rangle$ for a planning task $\Pi$, a successful block-substitution of a block $b_x\in \mathcal{B}$ with a block $\hat{b}_x$ yields a valid BDPO plan.
\end{theorem}
\begin{proofsketch}
Given $\pi_{bdp}$ is valid, every block precondition in $\pi_{bdp}$ is supported by a causal link with no threat, as proven in \cite{siddiqui_patrik_2012}.

A successful block-substitution of $b_x$ with $\hat{b}_x$, results in a BDPO plan $\pi'_{bdp}=\langle \mathcal{O'},\mathcal{B'} \prec'\rangle$ where all required causal links for the substituting block $\hat{b}_x$ are established, and the causal links, supported by the original block $b_x$, are re-established using $\hat{b}_x$ as producer. 

We will prove that no causal link in $\pi'_{bdp}$ has a threat. Let us first consider the causal links $b_i \xrightarrow{\langle v,d \rangle} b_j$ in $\pi'_{bdp}$  that are preexisted in $\pi_{bdp}$, i.e., $b_i, b_j \in \mathcal{B} \cap \mathcal{B}'$. No block $b\neq \hat{b}_x$ threatens $b_i \xrightarrow{\langle v,d \rangle} b_j$ in $\pi'_{bdp}$. If such a block $b$ were to threaten this causal link, it would also threaten $b_i \xrightarrow{\langle v,d \rangle} b_j$ in $\pi_{bdp}$. This cannot be true since $\pi_{bdp}$ is valid.  Given that the block-substitution is successful, the possible threats posed by $\hat{b}_x$ are resolved using the threat-resolving strategies.  
 
 Now, let us consider the causal links that are associated with $\hat{b}_x$. The causal links of the form $\hat{b}_x \xrightarrow{\langle v, d \rangle} b_j$ in $\pi'_{bdp}$ do not have any threat. If such a threat were present, the corresponding causal link $b_x \xrightarrow{\langle v, d \rangle} b_j$ in the original plan $\pi_{bdp}$ would also be threatened. This contradicts our initial assumption as $\pi_{bdp}$ is valid. Lastly, all threats to the causal links of the form $b_i \xrightarrow{\langle v, d \rangle} \hat{b}_x$ in $\pi'_{bdp}$ are resolved since the block-substitution is successful. Therefore, $\pi'_{bdp}$ is a valid BDPO plan, as each block precondition in $\pi'_{bdp}$ is supported by a causal link with no threat.
\end{proofsketch}

We use block substitution to replace one resource with another within blocks in BDPO plans, facilitating concurrent execution of the blocks that were previously not allowed to execute in parallel.

\subsection{Parallel Plan}
 Bäckström formalizes parallel plans by including an extra relation in partial-order plans \cite{backstrom1998}. This relation specifies which unordered operators cannot be executed concurrently. Following Bäckström, we give the definition of parallel plans.
\begin{definition}\label{def:parallel_plan}
A \textbf{parallel plan} is a triple $\pi_\mathcal{P}=\langle \mathcal{O}, \prec, \#\rangle$ for a planning problem $\Pi$, where $\langle \mathcal{O}, \prec\rangle$ is a partial-order plan, and $\#$ is an irreflexive, symmetric relation on $\mathcal{O}$ w.r.t. $\Pi$. A \emph{definite} plan is a parallel plan $\pi_p= \langle \mathcal{O}, \prec, \#\rangle$ such that $\#\subseteq (\prec \cup \prec^{-1})$. 
\end{definition}

The relation, denoted by \#, represents non-concurrency constraints between the operators w.r.t. a planning problem. Here, $o_i \prec^{-1} o_j$ indicates $o_j \prec o_i$ for two operators $o_i$ and $o_j$ \cite{backstrom1998}. A non-concurrency constraint between a pair of operators, $o_i$ and $o_j$, written as $o_i\#o_j$, states that $o_i$ and $o_j$ can not be executed in parallel. The relation \# is irreflexive, meaning that an operator can not have a non-concurrency constraint with itself.  It is also symmetric,  implying that if $o_i\#o_j$ is true for two operators $o_i$ and $o_j$, then $o_j\#o_i$ also holds.  Notably, \# specifies whether an operator pair can be executed in parallel w.r.t. to a planning problem. Even though an ordering constraint between two operators in a plan implicitly implies a non-concurrency constraint between them, it does not necessarily mean that the operators can not be executed in parallel under different circumstances \cite{backstrom1998}. Hence, $o_i\prec o_j$ does not necessarily imply $o_i\#o_j$.

A \emph{definite} parallel plan is a parallel plan $\pi_\mathcal{P}=\langle \mathcal{O}, \prec, \#\rangle$ where there is no non-concurrency constraint in any pair of unordered operators (i.e. $\#\subseteq (\prec \cup \prec^{-1})$). In other words, POPs in which all unordered operators are parallel are definite plans. 
Bäckström \cite{backstrom1998} provides some conditions for a non-concurrency constraint between two operators, based on the facts that they consume, produce, and delete (Definition \ref{def:op}).  
According to Bäckström, a non-concurrency constraint between two operators $o_i$ and $o_j$, $o_i\#o_j$, holds if any of the following conditions is true for a fact $f$.

\begin{enumerate}
    \item $f\in (prod(o_i) \cap cons(o_j))$.
    \item $f\in (prod(o_i) \cap del(o_j))$.
    \item $f\in (cons(o_i) \cap del(o_j))$.
\end{enumerate}

The first condition signifies the $PC$ reason, while the second and the third conditions indicate $DP$ and $CD$ reasons, respectively (Definition \ref{def:pc-cd-dp}). Notably, Bäckström mentions that two producers, two consumers, or two deleters of the same facts are not always required to be non-concurrent for a plan with multiple producers for that fact \cite{backstrom1998}.  He also points out that it is entirely permissible to impose additional conditions for non-concurrency constraints between operators of a plan. While these conditions can be sufficient to identify non-concurrency constraints between two operators within a plan, they are not always sufficient w.r.t. a planning problem. Let us illustrate with an example.
    
Let us recall the planning task from the Elevator domain, presented in Example \ref{example:sas},  where there are two lifts \verb|e1| and \verb|e2|, two passengers \verb|p1| and \verb|p2|, and three floors \verb|n1|, \verb|n2| and \verb|n3|. In this domain, operators \verb|board p1 n1 e1| and \verb|board p2 n2 e1| board passengers \verb|p1| and \verb|p2| on the same lift \verb|e1| from floors \verb|n1| and \verb|n2|, respectively. The operator \verb|board p1 n1 e1| consumes the facts $\langle v_{e1}, n1\rangle$ and $\langle v_{p1}, n1\rangle$, and produces $\langle v_{p1}, e1\rangle$. Therefore, this operator deletes $\langle v_{p1}, n1\rangle$. On the other hand, Operator \verb|board p2 n2 e1| consumes $\langle v_{e1}, n2\rangle$ and $\langle v_{p2},n2\rangle$, produces $\langle v_{p2},e1\rangle$, and deletes $\langle v_{p2}, n2\rangle$.  These two operators do not consume. produce or delete any common fact. According to the conditions by Bäckström \cite{backstorm1994}, these operators \verb|board p1 n1 e1| and \verb|board p2 n2 e1| do not have any non-concurrency constraint. However, these operators are not concurrent, as the same lift \verb|e1| cannot be on different floors simultaneously.  The facts in the preconditions or the effects of these operators can not recognize that these operators are using the same resource (i.e., lift \verb|e1|).  Note that a valid POP will never allow these operators to be executed in parallel, as they will be ordered due to at least a causal link, along with a promotion or demotion ordering. However, a non-concurrency constraint between two operators is defined for a planning problem independently of their ordering constraints, and this notion of non-concurrency constraint is necessary for our proposed methodology (presented in Section \ref{alg:cibs}). In this work, we redefine the necessary and sufficient conditions for a non-concurrency constraint between an operator pair in terms of state variables.

\subsection{Previous Approaches}
A parallel plan, similar to a partial-order plan, represents one or more sequential plans, but it does not necessarily imply temporal parallelism \cite{RINTANEN20061031}. Different notions of parallel plans can be defined. Knoblock classifies parallel plans into several categories \cite{Knoblock1994}. The most general class of parallel plans includes those where the parallel operators interact in some specific ways. Two operators are interacting if and only if they need to be executed in parallel or overlap in a specific manner for the plan to succeed. We focus on a more restricted class of parallel plans that do not involve interacting operators. In this class of parallel plans, the effects of executing the operators in any order are identical to the effects of executing the operators concurrently with respect to the goal for the corresponding planning task. Knoblock defines these operators as \emph{independent relative to a goal} \cite{Knoblock1994}. This study adopts this class of parallel plans to exploit potential parallelism and reduce the overall plan execution duration, not because the planning problem requires parallelism for the plan to succeed. 

To execute unordered operators or subplans (i.e., blocks of operators)  in parallel, they must have no resource conflict (i.e., interference). Researchers have proposed various approaches in the literature to address resource conflicts (i.e., operator interference) in a parallel plan. Regnier and Fade \cite{regnier_fade_1991, fade1991determination} presented an algorithm for converting a sequential plan into a parallel plan, in which all unordered operators are allowed to execute in parallel. This algorithm is a special case of an algorithm proposed earlier by Pednault \cite{PEDNAULT198747}. These algorithms add orderings between those unordered operators that cannot be executed in parallel, resulting in an over-constrained POP.  To resolve this issue, Bäckström \cite{backstorm1994} distinguishes parallel execution from operator unorderedness by incorporating non-concurrency constraints alongside the ordering constraints in a parallel plan. As described in the previous section, Bäckström also provides necessary conditions for a non-concurrency constraint between two operators in a plan. While these conditions are sufficient to establish a non-concurrency constraint within a given POP, they are not sufficient w.r.t. a planning problem. Moreover, they do not suffice to determine non-concurrency constraints between two subplans.

Another common approach is to explicitly model resource conflicts  within the problem descriptions. Knoblock incorporates a list in the \acrshort*{pddl} \cite{lipovetzky2019introduction} action definition to describe the resources utilized by an operator \cite{Knoblock1994}.  This resource list represents non-concurrency constraints, operators that employ the same resource cannot be executed at the same time. Then, their work modifies the UCPOP planner \cite{ucpop} to enforce these resource constraints. Boutilier and Brafman \cite{Boutilier_1997, Boutilier_2001}  address partial-order planning with concurrent interacting operators by including a \emph{concurrent operator list} in the operator specification. This list models which operators must be executed and which must not be executed concurrently with a particular operator. They also propose a partial-order planner capable of integrating such operators. Though their proposed language has several limitations \cite{kovacs2012multi}, their work demonstrates a way to model multi-agent planning with minimal changes to a single-agent base language (STRIPS). Their representation has been later extended to several applications of multi-agent planning \cite{SHEKHAR2020103200}. 

Since two ordered operators cannot be executed in parallel within a given plan, minimizing ordering constraints is crucial for enhancing the plan's concurrency. That is why, plan deordering and reordering techniques are also related to this study. Traditional plan deordering techniques, such as Explanation-based Order Generalization (EOG) \cite{kk, Veloso2002} and block deordering \cite{siddiqui_patrik_2012}, can efficiently deorder a plan in polynomial time, but they lack optimality guarantees. This limitation prompts MaxSAT-based reorderings \cite{maxsat} to encode the optimization of a POP’s orderings as a partial weighted MaxSAT problem. However, these reordering techniques become infeasible for large-scale plans due to their computational complexity \cite{maxsat}. In addition, a plan's ordering can be minimized by leveraging operators that are not initially included in the plan for a given planning problem. Previous deordering strategies, such as EOG and block deordering, do not alter the given plan in terms of operators. MaxSAT encodings by Muise et al. \cite{maxsat} allow eliminating redundant operators from the plan but do not permit introducing new ones. While Waters et al. \cite{maxsat_reinst, Waters_Nebel_Padgham_Sardina_2018} enables the replacement of operators by reinstantiating their parameters in their MaxSAT encodings, it comes with significant computational costs. Additionally, this encoding has some limitations, including the inability to substitute one operator with another with a different name. Noor and Siddiqui \cite{MaxSaT_sabah} combine iterative deordering approaches with these MaxSAT-based reorderings by employing block deordering on the MaxSAT solutions. They also introduce block-substitution to enable substituting blocks in a BDPO plan for a planning problem, and exploit block-substitution to further elevate the flexibility of a BDPO plan \cite{noor2024improving}. As shown in Example \ref{ex:example1}, though block deordering can often remove ordering by forming blocks, non-concurrency constraints between the blocks often persist.  In this work, we exploit block substitution to eliminate such non-concurrency constraints, further improving concurrency of a parallel plan.

\section{Parallel Block Decomposed Plan}\label{sec:parallel plan}
This work introduces the necessary semantics for enabling parallelism in BDPO plans. Similar to a parallel plan (Definition \ref{def:parallel_plan}), we facilitate execution concurrency in a BDPO plan by incorporating non-concurrency constraints over plan operators, allowing two unordered blocks with no non-concurrency constraint to be executed in parallel. This plan is referred to as \emph{Parallel block decomposed (PBD) plan}.

\begin{definition}
A \textbf{parallel block decomposed (PBD) plan} is a quadruple $\pi_{pbd}=\langle \mathcal{O}, \mathcal{B} \prec, \#\rangle$, where $\langle \mathcal{O}, \mathcal{B}, \prec\rangle$ is a BDPO plan and $\#$ is a  irreflexive, symmetric relation on $\mathcal{O}$, representing non-concurrency constraints between operator pairs.
\end{definition}

Two blocks $b_i$ and $b_j$ in a PBD plan have a non-concurrency constraint if an operator (i.e., primitive block) $o_i$ in block $b_i$ has a non-concurrency constraint with another operator $o_j$ in $b_j$. The set of variables for which a non-concurrency constraint holds between two blocks $b_i$ and $b_j$ is denoted as $vars_{nc}(b_i\#b_j)$.

\begin{definition}
    Let $\pi_{pbd}=\langle \mathcal{O}, \mathcal{B} \prec, \#\rangle$ be a  PBD plan, and  $b_i, b_j\in \mathcal{B}$ be two disjoint blocks. Blocks $b_i$ and $b_j$ have a non-concurrency constraint $b_i\# b_j$ if there are two operators $o_i\in b_i$ and $o_j\in b_j$ such that $o_i\# o_j$.
\end{definition}

\begin{figure}[!tbp]
    \centering
    \small
        \begin{tikzpicture}[node distance = 1cm]
               \node (init)[process] {\verb|INIT|};
                \node (op)[block2, below of = init, xshift = -2cm, yshift=-1.55cm]{$b_1$};
                \node (op1)[block2, below of = init, xshift = 2cm, yshift=-.6cm]{$b_2$};
                \node (op2)[block2, below of = op1]{$b_3$};
                \node (op3)[block2, below of = op2]{$b_4$};
            
                \node (goal)[process, below of= op3,  xshift =-2cm, yshift=-.4cm] {\verb|GOAL|};
                 \draw [arrow] (init) -> node[left]{$PC(\langle v_1, d_1\rangle)$}(op);
    
                \draw [arrow] (op1) ->node[left]{$PC(\langle v_1, d_1'\rangle)$}(op2); 
                \draw [arrow] (op2) ->node[left]{$PC(\langle v_1, d_1\rangle)$}(op3); 
                \draw [arrow, dashed] (op) ->(goal);  
            
                 
                 \begin{pgfonlayer}{background}
                     \node (fit1) [fit=(op1) (op2) (op3)] [block, inner xsep=27pt, xshift=-.6cm, label={[xshift=5.5mm]left:{$\bm{b}$}}] {};
                \end{pgfonlayer}
                    \draw [arrow, dashed] (fit1.south) ->(goal); 
                    \draw [arrow] (init) ->node[right]{$PC(\langle v_1, d_1\rangle)$} (fit1.north);
                     \draw[ultra thick] (op) --node[yshift=.3cm]{$\#$}(fit1);
            \end{tikzpicture}
            \caption{Blocks $b_1$ and $b$ had a non-concurrency constraint on a variable $v_1$ (i.e., $vars_{nc}=\{v_1\}$) since blocks $b_2$ and $b_3$ in $b$ have non-concurrency constraints with $b_1$ on the variable $v_1$.}
            \label{fig:concurrent_block}
    \end{figure}

We define the non-concurrency constraint between two blocks based on the primitive blocks (i.e., operators) they contain.  This is because if some primitive blocks use and release a resource within a block, the precondition and effects of that block can not capture the use of that resource. Let us consider the blocks $b_1$ and $b$ in the PBD plan presented in Figure \ref{fig:concurrent_block} and assume blocks $b_2, b_3, b_4 \in b$  are primitive blocks.  Both blocks $b_1$ and $b$ have precondition $\langle v_1, d_1\rangle$, and do not produce any fact of variable $v_1$. Block $b_2\in b$ consumes the fact $\langle v_1, d_1\rangle$, and changes the value of $v_1$ to $d_1'$. Then block $b_3\in b$ changes the value of $v_1$ back to $d_1$, thus nullifying the overall effect of $b_2$ on variable $v_1$ within the block $b$. However, there are conflicts over the resource associated with variable $v_1$ between blocks $b_1$ and $b_2$, and between $b_1$ and $ b_3$, causing the blocks $b_1$ and $b$ to be non-concurrent.

An ordering constraint between two blocks $b_i$ and $b_j$ in a PBD plan implicitly implies a non-concurrency constraint between them for that plan. In addition, if a block $b_i$ is contained in a block $b$, and blocks $b$ and $b_j$ have a non-concurrency constraint in a PBD plan, then it also implicitly implies $b_i\#b_j$ since two disjoint blocks can not interleave (Definition \ref{def:block1}). 
We refer to a non-concurrency constraint between a block pair as \emph{necessary non-concurrency constraint} when it is not implied by an ordering constraint or by a non-concurrency constraint involving blocks in which they are nested.


\begin{definition}
    Let $\pi_{pbd}=\langle \mathcal{O}, \mathcal{B} \prec, \#\rangle$ be a PBD plan, and $b_i\# b_j$ be a non-concurrency constraint between two disjoint blocks $b_i, b_j\in \mathcal{B}$. $b_i\# b_j$ is a \textbf{necessary non-concurrency constraint} w.r.t. $\pi_{pbd}$ if 1) $b_i\nprec b_j\nprec b_i$, and 2) if $b_i$ is in a block $b_x\in \mathcal{B}$, then $b_j$ is also in $b_x$.  
\end{definition}

To improve the concurrency of a PBD plan (as discussed later in Section \ref{sec:cibs}), we attempt to remove necessary non-concurrency constraints through block substituting. Because eliminating other non-concurrency constraints, such as non-concurrency constraints between ordered block pairs, does not contribute to improving the plan's concurrency.





\section{Conditions for Non-concurrency Constraints in a Parallel Plan} \label{sec:conditions}

A parallel plan specifies non-concurrency constraints over operators in a partial-order plan and allows two unordered operators with no non-concurrency constraint to be executed in parallel. 
FDR can capture relationships among objects using state variables, which group mutually exclusive propositions. 
Definition \ref{def:nonconcurrent operator} specifies the conditions for a non-concurrency constraint between an operator pair of a plan in terms of FDR variables. It is based on the principle that the value of a variable in a planning task cannot be different in the same state.

\begin{definition}\label{def:nonconcurrent operator}
    Let $\pi_\mathcal{P}=\langle \mathcal{O}, \prec, \#\rangle$, be a parallel plan for a planning task $\Pi= \langle \mathcal{V}, O, s_0, s_*\rangle$, and  $o_i, o_j\in \mathcal{O}$ be two operators. Operators $o_i$ and $o_j$ have a non-concurrency constraint, $o_i \# o_j$, if any of the following conditions holds for a variable  $v\in \mathcal{V}$,
    \begin{enumerate}
        \item  $pre(o_i)(v) \ne pre(o_j)(v)$.
         \item $\mathit{eff}(o_i)(v) \ne \mathit{eff}(o_j)(v)$.
         \item $pre(o_i)(v) \ne \mathit{eff}(o_j)(v)$.
    \end{enumerate}
\end{definition}

Definition \ref{def:nonconcurrent operator} states that two operators with different preconditions or effects on the same variable have a non-concurrency constraint.
 Let us clarify with a few examples. Operators \verb|move_up e1 n2 n3| and \verb|move_up e2 n2 n3| are concurrent since their precondition and effects have no common variable. Operators \verb|board p1 n1 e1| and \verb|board p2 n1 e1| are also concurrent, despite sharing a common variable $v_{e1}$ in their preconditions, because the value of $v_{e1}$ in both of their preconditions is $n1$. On the other hand, operators \verb|board p1 n1 e1| and \verb|board p2 n2 e1| are non-concurrent since they have different preconditions  \atom{$v_{e1}$}{n1} and \atom{$v_{e1}$}{n2}, respectively, on the same variable $v_{e1}$.  The operators \verb|move_up e1 n2 n3| and \verb|move_down e1 n2 n1| have a  non-concurrency constraint due to having different effects \atom{$v_{e1}$}{n3} and \atom{$v_{e1}$}{n1}, respectively, on $v_{e1}$. Operators \verb|board p1 n1 e1| and \verb|move_down e1 n1 n2| are also non-concurrent because \verb|move_down e1 n1 n2| has an effect that is different from the precondition of the operator \verb|board p1 n1 e1| involving the same variable.

\begin{theorem}
\sloppy{
    The conditions in Definition \ref{def:nonconcurrent operator} are necessary for a non-concurrency constraint between an operator pair in a POP.}
\end{theorem}
\begin{proofsketch}
    Let $\pi=\langle \mathcal{O}, \prec\rangle$ be partial-order plan and $o_i, o_j\in \mathcal{O}$ be two operators in $\pi$. We will prove that each condition in Definition \ref{def:nonconcurrent operator} is necessary for a non-concurrency constraint between the operator pair $o_i$ and $o_j$. 

    Let $\langle v, d \rangle$ and $\langle v, d' \rangle$ be two facts where  $d, d'\in \mathcal{D}_v$ and $d\ne d'$. These two facts belongs to the same variable $v$. Hence, they are mutually exclusive propositions and can not be true in the same state.
    
    Condition 1: Assume, for contradiction, that the first condition $pre(o_i)(v)\ne pre(o_j)(v)$ is not necessary. Let $\langle v, d \rangle$ be in  $pre(o_i)$ and $\langle v, d' \rangle$ be in $pre(o_j)$. If $o_i$ and $o_j$ are allowed to execute in parallel, then  $\langle v, d \rangle$ and $\langle v, d' \rangle$ must hold in a state, which can not be true. Therefore, the condition $pre(o_i)(v)\ne pre(o_j)(v)$ is necessary.
    
    Condition 2: Assume that the second condition is not necessary. Let us consider  $\langle v, d \rangle \in \mathit{eff}(o_i)$ and $\langle v, d' \rangle \in \mathit{eff}(o_j)$. If $o_i$ and $o_j$ are concurrent, then  $\langle v, d \rangle$ and $\langle v, d' \rangle$ must hold in a state, which can not be true. Therefore, the condition $\mathit{eff}(o_i)(v)\ne \mathit{eff}(o_j)(v)$ is necessary. 
    
    Condition 3: Assume that the third condition is not necessary. Let us consider  $\langle v, d \rangle \in pre(o_i)$ and $\langle v, d' \rangle \in \mathit{eff}(o_j)$.   If $o_i$ and $o_j$ are concurrent, the value of the variable $v$ is $d$ in the state in which $o_i$ and $o_j$ are applied. Therefore, $o_j$ deletes the fact $\langle v, d \rangle$ and becomes a threat to $o_i$, requiring an ordering constraint between $o_i$ and $o_j$ for the plan to be valid. Thus, the third condition $pre(o_i)(v)\ne \mathit{eff}(o_j)(v)$ is also necessary for a non-concurrency constraint between an operator pair.
\end{proofsketch}
\begin{theorem}
\sloppy{
    The conditions in Definition \ref{def:nonconcurrent operator} are sufficient for a non-concurrency constraint between an operator pair in a POP.}
\end{theorem}
\begin{proofsketch}
    Let $\pi=\langle \mathcal{O}, \prec\rangle$ be partial-order plan and $o_i, o_j\in \mathcal{O}$ be two operators in $\pi$.  We will prove that no condition other than those outlined in Definition \ref{def:nonconcurrent operator} is responsible for a non-concurrency constraint between the operator pair $o_i$ and $o_j$. There are three other possible conditions : (i) $pre(o_i)(v) = pre(o_j)(v)$, (ii) $\mathit{eff}(o_i)(v) =  \mathit{eff}(o_j)(v)$, and (iii) $pre(o_i)(v) = \mathit{eff}(o_j)(v)$. 

   First condition $pre(o_i)(v) = pre(o_j)(v)$: Let $\langle v, d \rangle$ be in both $pre(o_i)$ and $pre(o_j)$. To execute both $o_i$ and $o_j$ at the same time, the value of $v$ must be $d$ in the state where these operators are to be applied.  Since there is no conflict, this condition is not necessary for a non-concurrency constraint between $o_i$ and $o_j$. 

   Second condition $\mathit{eff}(o_i)(v) = \mathit{eff}(o_j)(v)$:  Let $\langle v, d \rangle$ be in  $\mathit{eff}(o_i)$ and $\langle v, d \rangle$ be in $\mathit{eff}(o_j)$. $\langle v, d \rangle \in \mathit{eff}(o_i)$ implies either $v \notin vars(pre(o_i))$ or $\langle v, d' \rangle \in pre(o_i)$, where $d'\in \mathcal{D}_v$, $d\ne d'$. Situation 1: when $v \notin vars(pre(o_i))$ and $v \notin vars(pre(o_j))$, there is no conflict since the value of $v$ becomes $d$  in the state, generated after executing both operators in parallel. Situation 2:  when  $\langle v, d' \rangle \in pre(o_i)$ or  $\langle v, d' \rangle \in pre(o_j)$ holds, then operator $o_i$ and $o_j$ are not concurrent due to the third condition in Definition \ref{def:nonconcurrent operator}. Hence, the condition  $\mathit{eff}(o_i)(v) = \mathit{eff}(o_j)(v)$ is not necessary.
 
   Third condition $pre(o_i)(v) = \mathit{eff}(o_j)(v)$:  Let the fact  $\langle v, d \rangle$ be in $pre(o_i)$ and $\mathit{eff}(o_j)$ in a POP. Situation 1: $o_j$ is responsible for providing $\langle v,d \rangle$ to $o_i$, thereby forming a causal link $o_j \xrightarrow{\langle v, d\rangle} o_i$ and an ordering  $o_j \prec o_i$. Then, operators $o_i$ and $o_j$ are non-concurrent, and this non-concurrency is implied by the ordering constraint  $o_j \prec o_i$ for this plan. Situation 2: There exists another operator $o_k$ that provides the fact $\langle v, d \rangle$ to $o_i$ through a causal link $o_k \xrightarrow{\langle v, d \rangle} o_i$. Since $o_j$ does not threaten $o_k \xrightarrow{\langle v, d \rangle} o_i$, the plan remains valid. Therefore, this condition $pre(o_i)(v) = \mathit{eff}(o_j)(v)$ is not necessary for a non-concurrency constraint between $o_i$ and $o_j$. 
\end{proofsketch}

\section{Minimizing Non-Concurrency Constraints in Parallel Plans}\label{sec:sec4}
This work transforms a parallel plan into a parallel block decomposed (PBD) plan using block deordering, and then exploits block substitution to further enhance the concurrency of that plan. 
Before applying block substitution to eliminate non-concurrency constraints between two blocks in a PBD plan, it is often necessary to extend a block by incorporating some of its predecessors or successors.

\subsection{Extending Blocks for Substitution}\label{sec:extend}
We can interpret a block as a macro-action, which has some preconditions and effects. A block also represents a partial-order subplan. Let us consider a block $b$ in a PBD plan $\pi_{pbd}=\langle \mathcal{O}, \mathcal{B}, \prec, \# \rangle$, and $\pi=\langle \hat{\mathcal{O}}, \hat{\prec}\rangle$ is the partial-order subplan represented by $b$.  Assume $b$ has a precondition $\langle v, d\rangle$, and it also provides a fact $\langle v, d'\rangle$ to some block in $\pi_{pbd}$  where $d\ne d'$. If the value of variable $v$ can not be transitioned from $d$ to $d'$ by a POP other than $\pi$, then the block $b$ can not be substituted by another block or subplan.  We can determine such situations using \acrfull*{dtg}. However, extending the block with some of its predecessors or successors can enable substitution. Let us illustrate such a situation with an example.

\begin{example}
    \tikzstyle{block3} = [draw, inner xsep= 5.5pt, color=black!80, fill=black!7, xshift=-.15cm]
\begin{figure}[!tbp]
\centering
    \begin{minipage}{.58\textwidth}
         \begin{subfigure}{\textwidth}
    \fontsize{8}{7}
    \begin{tikzpicture}[node distance = 1cm]
            \node (init)[process] {\verb|INIT|};
            \node (op1)[process, below of = init, yshift=-.4cm, xshift=-1.7cm]{\verb|board p1 n2 e1|};
            \node (op2)[process, below of = op1, yshift=-.2cm, xshift=-.3cm]{\verb|move_up e1 n2 n3|};
            \node (op3)[process, below of = op2]{\verb|leave p1 n3 e1|};
            \node (op4)[process, below of = op3]{\verb|move_down e1 n3 n2|};
             \node (op5)[process, below of = init, xshift=2cm, yshift=-.4cm]{\verb|move_down e1 n2 n1|};
             \node (op6)[process, below of = op5]{\verb|board p2 n1 e1|};
             \node (op7)[process, below of = op6]{\verb|move_up e1 n1 n2|};
            \node (op8)[process, below of = op7, yshift=-.2cm]{\verb|leave p2 n2 e1|};
            \node (goal)[process, below of= op8, xshift=-2cm, yshift=-.3cm] {\verb|GOAL|\tikzmark{foo1}};
        \begin{pgfonlayer}{background}
            \node (b1) [fit=(op1)] [block3, label={[xshift=4.5mm]left:{$b_1$}}] {};
            \node (b2) [fit=(op2)] [block3, label={[xshift=4.5mm]left:{$b_2$}}] {};
            \node (b3) [fit=(op3)] [block3, label={[xshift=4.5mm]left:{$b_3$}}] {};
            \node (b4) [fit=(op4)] [block3, label={[xshift=4.5mm]left:{$b_4$}}] {};
            \node (b5) [fit=(op5)] [block3, label={[xshift=4.5mm]left:{$b_5$}}] {};
            \node (b6) [fit=(op6)] [block3, label={[xshift=4.5mm]left:{$b_6$}}] {};
            \node (b7) [fit=(op7)] [block3, label={[xshift=4.5mm]left:{$b_7$}}] {};
            \node (b8) [fit=(op8)] [block3, label={[xshift=4.5mm]left:{$b_8$}}] {};
             \node (fit1) [fit=(b2) (b3) (b4)] [block, xshift =.2cm,inner xsep= 2.5pt, fill=none, label={[xshift=4.2mm]left:{\footnotesize{$\mathbf{b_i}$}}}] {};
             \node (fit2) [fit=(b5) (b6) (b7)] [block, , xshift =.2cm,inner xsep= 2.5pt, fill=none, label={[xshift=4.2mm]left:{\footnotesize{$\mathbf{b_j}$}}}]{};
        \end{pgfonlayer}
            \draw [arrow] (init) -> node[anchor=south east]{PC(\atom{$v_{e1}$}{2})}node[anchor= east]{PC(\atom{$v_{p1}$}{n2})}(b1);
            \draw [arrow] (b1) -> node[anchor=east]{PC(\atom{$v_{p1}$}{e1})}(fit1);
            \draw [arrow] (b2) -> (b3);
            \draw [arrow] (b3) -> (b4);
           \draw [arrow] (b5) -> (b6);
            \draw [arrow] (b6) ->(b7);
           \draw [arrow] (fit2) ->node[anchor= east]{PC(\atom{$v_{e1}$}{n2})}(b8);
            \draw [arrow] (b8) ->node[anchor= west]{PC(\atom{$v_{p2}$}{n2})}(goal); 
            \draw [arrow] (fit1) -> node[anchor= east]{PC(\atom{$v_{p1}$}{n3})}(goal);
            \draw [arrow] (init) -> node[anchor= south west]{PC(\atom{$v_{e1}$}{n2})}node[anchor= west]{PC(\atom{$v_{p2}$}{n2})}(fit2);
            \draw[ultra thick] (fit1) -- node[yshift=.3cm]{$\#$}(fit2);
        \end{tikzpicture}  
        \caption{}
    \end{subfigure}
    \begin{subfigure}{\textwidth}
    \fontsize{8}{7}
    \begin{tikzpicture}[node distance = 1cm]
            \node (init0)[rectangle]{};
            \node (init)[process, below of = init0, yshift=.5cm] {\verb|INIT|};
            \node (op1)[process, below of = init, yshift=-.4cm, xshift=-2cm]{\verb|board p1 n2 e1|};
            \node (op2)[process, below of = op1, yshift=-.2cm,]{\verb|move_up e1 n2 n3|};
            \node (op3)[process, below of = op2]{\verb|leave p1 n3 e1|};
            \node (op4)[process, below of = op3]{\verb|move_down e1 n3 n2|};
             \node (op5)[process, below of = init, xshift=2cm, yshift=-.4cm]{\verb|move_down e1 n2 n1|};
             \node (op6)[process, below of = op5]{\verb|board p2 n1 e1|};
             \node (op7)[process, below of = op6]{\verb|move_up e1 n1 n2|};
            \node (op8)[process, below of = op7, yshift=-.2cm]{\verb|leave p2 n2 e1|};
            \node (goal)[process, below of= op8, xshift=-2cm, yshift=-.3cm] {\verb|GOAL|};

        \begin{pgfonlayer}{background}
            \node (b1) [fit=(op1)] [block3, label={[xshift=4.5mm]left:{$b_1$}}] {};
            \node (b2) [fit=(op2)] [block3, label={[xshift=4.5mm]left:{$b_2$}}] {};
            \node (b3) [fit=(op3)] [block3, label={[xshift=4.5mm]left:{$b_3$}}] {};
            \node (b4) [fit=(op4)] [block3, label={[xshift=4.5mm]left:{$b_4$}}] {};
            \node (b5) [fit=(op5)] [block3, label={[xshift=4.5mm]left:{$b_5$}}] {};
            \node (b6) [fit=(op6)] [block3, label={[xshift=4.5mm]left:{$b_6$}}] {};
            \node (b7) [fit=(op7)] [block3, label={[xshift=4.5mm]left:{$b_7$}}] {};
            \node (b8) [fit=(op8)] [block3, label={[xshift=4.5mm]left:{$b_8$}}] {};
             \node (fit1) [fit=(b1)(b2) (b3) (b4)] [block,ultra thick, xshift =.2cm,inner xsep= 2.5pt, fill=none, label={[xshift=4.8mm, yshift=-1.8mm]left:{\small{$\mathbf{b_i'}$}}}] {};
             \node (fit2) [fit=(b5) (b6) (b7)] [block, , xshift =.2cm,inner xsep= 2.2pt, fill=none, label={[xshift=4.2mm]left:{\footnotesize{$\mathbf{b_j}$}}}]{};
        \end{pgfonlayer}
            
            \draw [arrow] (init) -> node[anchor=south east]{PC(\atom{$v_{e1}$}{2})}node[anchor= east]{PC(\atom{$v_{p1}$}{n2})}(fit1);
            \draw [arrow] (b1) -> (b2);
            \draw [arrow] (b2) -> (b3);
            \draw [arrow] (b3) -> (b4);
           
           \draw [arrow] (b5) -> (b6);
            \draw [arrow] (b6) ->(b7);
           \draw [arrow] (fit2) ->node[anchor= east]{PC(\atom{$v_{e1}$}{n2})}(b8);
            \draw [arrow] (b8) ->node[anchor= west]{PC(\atom{$v_{p2}$}{n2})}(goal);
        
        \draw [arrow] (fit1) -> node[anchor= east]{PC(\atom{$v_{p1}$}{n3})}(goal);
        \draw [arrow] (init) -> node[anchor= south west]{PC(\atom{$v_{e1}$}{n2})}node[anchor= west]{PC(\atom{$v_{p2}$}{n2})}(fit2);
        \draw[ultra thick] (fit1) -- node[yshift=.3cm]{$\#$}(fit2);
        \end{tikzpicture}
    \caption{}
    \end{subfigure}
    \end{minipage}
    \begin{minipage}{.4\textwidth}
        \begin{subfigure}[c]{\textwidth}
      \footnotesize{
          \begin{tikzpicture}[node distance = 4.1cm]
                \node (d1)[process] {$n_1$};
                \node (d2)[process, right of=d1] {$e1$};
                \node (d3)[process, below of=d2]{$n_2$};
                \node (d4)[process, left of=d3]{$e_2$};
                \node (d5)[process, left of=d3, yshift=2.05cm, xshift=2.05cm]{$n_3$};
                
                \draw [arrow] (d1.20) ->node[yshift =.2cm]{board p1 n1 e1}(d2.160);
                \draw [arrow] (d2.200) ->node[yshift =-.2cm]{leave p1 n1 e1}(d1.340);
                \draw [arrow] (d3.110) ->node[sloped, anchor=center, above]{board p1 n2 e1}(d2.250);
                \draw [arrow] (d2.290) ->node[sloped, anchor=center, above]{leave p1 n2 e1}(d3.70);
                
                \draw [arrow] (d4.20) ->node[sloped, anchor=center, above]{leave p1 n2 e2}(d3.160);
                \draw [arrow] (d3.200) ->node[sloped, anchor=center, below]{board p1 n2 e2}(d4.340);
                  \draw [arrow] (d4.110) ->node[sloped, anchor=center, above]{leave p1 n1 e2}(d1.250);
                \draw [arrow] (d1.290) ->node[sloped, anchor=center, above]{board p1 n1 e2}(d4.70);

                \draw [arrow] (d4.25) ->node[sloped, anchor=center, below]{leave p1 n3 e2}(d5.245);
                \draw [arrow] (d5.205) ->node[sloped, anchor=center, above]{board p1 n3 e2}(d4.65);
                \draw [arrow] (d5.25) ->node[sloped, anchor=center, below]{board p1 n3 e1}(d2.245);
                \draw [arrow] (d2.205) ->node[sloped, anchor=center, above]{leave p1 n3 e1}(d5.65);
            \end{tikzpicture}     
            }
    \caption{}
    \end{subfigure}
    \end{minipage}

    \caption{(a) Blocks $b_i$ and $b_j$ have a non-concurrency constraint, where $b_i$ changes the value of the variable $v_{p1}$ from $e1$ to $n2$, (b)  Extending the block $b_i$ to $b_i'$ by including block $b_1$, enabling the substitution of the block $b_i'$, (c) \acrfull*{dtg} of the variable $v_{p1}$.}
    \label{fig:extend}
\end{figure}
    Figure \ref{fig:extend}(a) illustrates a \acrfull*{pbd} plan corresponding to the BDPO plan, presented in Example \ref{example:bd}, from the Elevator domain. Blocks $b_i$ and $b_j$ in this PBD plan have a non-concurrency constraint because they share the same resource, lift $e1$.
    The block $b_i$ contains three operators that move the lift $e1$ up from floor $n2$ to $n3$, dropping off the passenger at floor $n3$, and then move lift $e1$ down from $n3$ to $n1$.  If we want to substitute the block $b_i$ with another block $\hat{b}_i$ that has no non-concurrency constraint with $b_j$, then $\hat{b}_i$ must use a different lift than $e1$. 
    However, the responsible operator for placing $p_1$ in the lift $e1$ is contained in block $b_1$, which is a predecessor of $b_i$. Since the passenger $p_1$ is already placed in lift $e1$ by block $b_1$ before block $b_i$, substituting $b_i$ is not possible. However, if we extend the block $b_i$ by including $b_1$, then the new extended block can be substituted by another block that does not use the lift $e1$.
    
    This scenario is also evident from the \acrfull*{dtg} of the variable $v_{p1}$, $G_{v_{p1}}$, presented in Figure \ref{fig:extend}(c). Block $b_i$ provides the fact $\langle v_{p1}, n2 \rangle$ to \verb|GOAL|, and has a precondition $\langle v_{p1}, e1 \rangle$ on the same variable $v_{p1}$. Therefore, the substituting block $\hat{b}_i$ must transition the value of the variable $v_{p1}$ from $e1$ to $n2$. DTG of $v_{p1}$ shows that this transition is not possible without operators using lift $e1$. In Figure \ref{fig:extend}(b), the block $b_i'$ is formed by extending $b_i$ up to $b_1$. The precondition of $b_i'$ becomes $\langle v_{p1}, n2\rangle$.  $G_{v_{p1}}$ in Figure \ref{fig:extend}(c) shows that there are two possible transition paths for the variable $v_{p1}$ from $n2$ to $n3$, one using $e1$ and another using $e2$. Therefore, the extended block $b_i'$ can be substituted by a block $\hat{b}_i$ that uses an alternate lift, yielding a PBD plan having no non-concurrency constraint between blocks $\hat{b}_i$ and $b_j$. 
\end{example}

 We present the EXTEND procedure in Algorithm \ref{alg:extend}, which takes two blocks $b_i$ and $b_j$ having a necessary non-concurrency constraint $b_i\#b_j$ in a PBD plan $\pi_{pbd}$ for a planning task $\Pi$. This procedure extends the block $b_i$ with necessary blocks so that the new extended block $b_i'$ can be substituted with a block $\hat{b}_i$ such that $vars_{nc}(\hat{b}_i\#b_j)=\emptyset$.
 
 EXTEND procedure (Algorithm \ref{alg:extend}) first calculates state $s$ by applying a subplan, generated by linearizing the predecessor blocks of $b_i$, to the initial state of $\Pi$ (line 3-4).  Then, it estimates the facts (denoted as $F$) that $b_i$ provides other blocks through causal links (line 5). It initializes an empty list $S$ to contain the blocks to be extended with block $b_i$.

 \begin{algorithm}[!t]
    \caption{Extending a Block}
    \label{alg:extend}
    \textbf{Input}:  A valid PBD plan $\pi_{pbd}=\langle \mathcal{O},\mathcal{B} \prec, \#\rangle$ for a planning task $\Pi= \langle \mathcal{V}, O, s_0, s_*\rangle$, and two blocks $b_i, b_j \in \mathcal{B}$ s.t. there is a necessary non-concurrency constraint   $b_i \# b_j$.\\
    \textbf{Output:} the block $b_i$, expanded with its relevant neighbors for substitution if necessary
    \begin{algorithmic}[1] 
    \Procedure{Extend}{$\Pi, \pi_{pbd}, b_i, b_j$} 
        \State $cvars \leftarrow vars_{nc}(b_i\#b_j)$
        \State  $\hat{\pi}\leftarrow$ get a sequential plan by linearizing  blocks $b\in \mathcal{B}$ s.t. $b_I \prec b \prec b_i$, and $b_I=\{o_I\}$ 
        \State $s \leftarrow$ $apply(\hat{\pi},s_0)$
        \State $F \leftarrow \{\langle v,d \rangle \mid b_i \xrightarrow{\langle v,d \rangle} b_k, b_k \in \mathcal{B}$\}
        \State $S = \emptyset$
        \ForAll{$b \in Immediate\_Predecessor(b_i)$ s.t $b\nprec b_j$}
            \ForAll{variable $v_1$ s.t $b\xrightarrow{\langle v_1, d_1\rangle}b_i, v_1\in vars(F)$}
                \State get transition path set $P$ in DTG $G_{v_1}$ from the value $d_1$ to $F(v_1)$
    
                \If{no path  $p\in P$ exists s.t. $\forall o\in p,\  vars_{nc}(o\#b_j)= \emptyset$ }
                \State $S \leftarrow S\cup b$
                \EndIf
            \EndFor
        \EndFor
        \If{$S$ is empty}
            \ForAll{$b \in Immediate\_Successor(b_i)$ s.t $b_j\nprec b$}
                \ForAll{variable $v_1$ s.t $b_i\xrightarrow{\langle v_1, d_1\rangle}b$}

                \State get transition path set $P$ in DTG $G_{v_1}$ from the value $s(v_1)$ to $d_1$
    
                \If{no path  $p\in P$ exists s.t. $\forall o\in p, \  vars_{nc}(o\#b_j)= \emptyset$ }
                \State $S \leftarrow S\cup b$
                \EndIf
                \EndFor
            \EndFor
        \EndIf
        \If{$S$ is empty}
            \Return $ b_i$
        \EndIf
        \State $b'_i = b_i \cup S$
        \State \Return \Call{Extend}{$\Pi, \pi_{pbd}, b'_i, b_j$}
    \EndProcedure
    \end{algorithmic}
\end{algorithm}

 This procedure iterates over each block $b$, which is an immediate predecessor of $b_i$ and is unordered with $b_j$, to determine if $b_i$ should be extended with $b$ (lines 7-14).  For each variable $v_1$ such that 1) there is a causal link $b\xrightarrow{\langle v_1, d_1\rangle} b_i$ and 2) $v_1$ is in $vars(F)$, this procedure first gets Domain Transition Graph  $G_{v_1}$ for the variable $v_1$. Then, it extracts the transition path set $P$ from value  $d_1$ to $F(v_1)$ in $G_{v_1}$. If there is no path $p\in P$, in which no operator has any non-concurrency constraint with $b_j$, then we add $b$ to $S$. 

 If $S$ is empty, this procedure iterates through each block $b$ that immediately follows $b_i$, and has no ordering with $b_j$, checking whether  $b_i$ needs to be extended with $b$ (lines 15-24).
 Similarly, for every  variable $v_1$ with a causal link $b_i\xrightarrow{\langle v_1, d_1\rangle}b$,  the procedure gets the transition path set $P$ in $G_{v_1}$ from $s(v)$ to $d_1$. If there exists no path in $P$ that does not have any operator with a non-concurrency constraint with $b_j$, then we add $b$  to $S$. The block $b_i$ is then extended with the blocks in $S$. The extended block is denoted as $b_i'$. This procedure continues to extend $b_i'$ recursively until $S$ is empty in a single iteration.

 In the following section, we describe our algorithm for improving concurrency in a PBD plan by substituting blocks. Our algorithm employs the EXTEND procedure to perform the necessary extension of a block before substituting it.

\subsection{ Concurrency Improvement via Block-Substitution} \label{sec:cibs}
We present an algorithm named \acrshort*{cibs} (Concurrency Improvement via Block-Substitution) to minimize non-concurrency constraints in a PBD plan using block-substitution. 

\begin{algorithm}[!tb]
    \caption{ Concurrency Improvement via Block-Substitution (CIBS)}
    \label{alg:cibs}
    \textbf{Input}: A valid sequential plan $\pi$ w.r.t. a planning task $\Pi$.\\
    \textbf{Output:} A valid PBD plan
    \begin{algorithmic}[1] 
    \State $\pi_{eog} \equiv \langle \mathcal{O}, \prec, \# \rangle \leftarrow$ \Call{EOG}{$\pi$}
    \Comment{generate parallel plan using EOG}
    \State $\pi_{bd} \equiv \langle \mathcal{O},\mathcal{B}, \prec, \# \rangle \leftarrow$ \Call{Block-Deorder} {$\pi_{eog}$} 
    \Comment{build compound blocks}
    \State $\pi_{sc}\leftarrow$ \Call{Substitution-For-Concurrency}{$\Pi, \pi_{bd}$} 
    \State \Return $\pi_{sc}$

\Procedure{Substitution-For-Concurrency}{$\Pi, \pi_{pbd}$}
        \ForAll{ necessary non-concurrency constraint $(b_i \# b_j) \in \#$}
            \State ($\hat{\pi}_{pbd}, success$) $\leftarrow$ \Call{Resolve\_Nonconcurrency}{$\Pi, \pi_{pbd}, b_i, b_j$}
            \Comment{try substituting $b_i$}
            \If{$success$ is \FALSE}
                \State ($\hat{\pi}_{pbd}, success$) $\leftarrow$ \Call{Resolve\_Nonconcurrency}{$\Pi, \pi_{pbd}, b_j, b_i$}
               \Comment{try substituting $b_j$}
            \EndIf
            \If{$success$} 
            \State \Return \Call{Substitution-For-Concurrency}{$\Pi, \hat{\pi}_{pbd}$}
            \EndIf
        \EndFor
        \State \Return $\pi_{pbd}$
    \EndProcedure
    \end{algorithmic}
\end{algorithm}

\acrshort*{cibs} (Algorithm \ref{alg:cibs}) takes a valid sequential plan $\pi$ for a planning task $\Pi$ as input, and produces a valid PBD plan, while enhancing the concurrency in three phases. The first phase converts $\pi$ into a parallel plan $\pi_{eog}= \langle \mathcal{O}, \prec,\#\rangle$ by first applying \acrshort*{eog}, and then identifying every non-concurrency constraint between each operator pair. Then, \acrfull*{bd} phase eliminates orderings from $\pi_{eog}$ by encapsulating cohesive operators in blocks, resulting in a PBD plan $\pi_{bd} = \langle \mathcal{O}, \mathcal{B}, \prec, \#\rangle$. Finally, the SUBSTITUTION-FOR-CONCURRENCY (SC) phase replaces blocks to further improve the concurrency of $\pi_{bd}$.

\sloppy{
The SUBSTITUTION-FOR-CONCURRENCY procedure, as outlined in Algorithm \ref{alg:cibs}, takes a PBD plan $\pi_{pbd} = \langle \mathcal{O}, \mathcal{B}, \prec, \#\rangle$ as input.
and iterates through each necessary non-concurrency constraint $b_i\#b_j$ between two blocks $b_i, b_j \in \mathcal{B}$, from the beginning of the plan.
 Then it attempts to remove $b_i\#b_j$ first by substituting $b_i$, then if this attempt fails, the procedure attempts to substitute $b_j$. After successfully removing the non-concurrency constraint between a block pair, it restarts from the beginning of the latest PBD plan. This process continues until no non-concurrency constraint can be successfully removed after a complete examination of all non-concurrency constraints in the plan. This procedure employs the RESOLVE\_NONCONCURRENCY procedure (Algorithm \ref{alg:resolve_concurrency}) to eliminate the non-concurrency constraint between a block pair.} 

\begin{algorithm}[!tbp]
 \caption{Resolve non-concurrency constraint between a pair of blocks via block-substitution}
    \label{alg:resolve_concurrency}
    \textbf{Input}:  A valid PBD plan $\pi_{pbd}=\langle \mathcal{O},\mathcal{B} \prec, \#\rangle$ for a planning task $\Pi= \langle \mathcal{V}, O, s_0, s_*\rangle$, and two blocks $b_i, b_j \in \mathcal{B}$ s.t. there is a necessary non-concurrency constraint   $b_i \# b_j$.\\
    \textbf{Output:} A PBD plan, and a boolean value indicating whether the targeted non-concurrency constraint is resolved
    \begin{algorithmic}[1] 
    \Procedure{Resolve\_Nonconcurrency}{$\Pi, \pi_{pbd}, b_i, b_j$}
        \State $b_i' = $\Call{Extend}{$\Pi, \pi_{pbd}, b_i, b_j$}
        \State  $\hat{\pi}\leftarrow$ linearize blocks $b\in \mathcal{B}$ s.t. $b_I \prec b \prec b_i'$, and $b_I=\{o_I\}$ 
        \State $\hat{s}_0 \leftarrow$ $apply(\hat{\pi},s_0)$ \Comment{get initial state for subtask}
        \State $G \leftarrow \{\langle v,d \rangle \mid b_i' \xrightarrow{\langle v,d \rangle} b_k, b_k \in \mathcal{B}$\}
         \State $C \leftarrow \{\langle v,d \rangle \mid b_x \xrightarrow{\langle v,d \rangle} b_y$, $b_x \prec b_i' \prec b_y,\{b_x, b_y\}\in \mathcal{B}\}$ 
          \State  $\hat{s}_* \leftarrow G \cup C$ \Comment{get goal for subtask}
        \State construct a subtask $\Pi_{sub}=\langle \mathcal{V}, O,\hat{s}_0$, $\hat{s}_*\rangle$ 
            \State $subplans$ $\leftarrow$  \Call {generate\_plans}{$\Pi_{sub} $} 
            \ForAll{$\hat{\pi} \in subplans$}
                \State $\hat{\pi}_{pop} \leftarrow$ apply EOG on $\hat{\pi}$
                \State make a new block $\hat{b}_i$ from $\hat{\pi}_{pop}$ 
                \If{$vars_{nc}(\hat{b}_i \# b_j) = \emptyset$} 
                \Comment{$\hat{b}_i$ and $b_j$ are concurrent}
                    \State ($\hat{\pi}_{pbd}, success$) $\leftarrow$  \Call{Substitute}{$\pi_{pbd}$, $b_i', \hat{b}_i$}
                    \If{ $success$ \textbf{and} $\mathit{cflex}(\hat{\pi}_{pbd}) > \mathit{cflex}(\pi_{pbd})$ \textbf{and} $cost(\hat{\pi}_{pbd}) \leq cost(\pi_{pbd})$}
                        \State \Return $\hat{\pi}_{pbd}$, \TRUE
                    \EndIf
                \EndIf
            \EndFor
        \State\Return  $\pi_{pbd}$, \FALSE
    \EndProcedure
    \end{algorithmic}
\end{algorithm}

The central component of the \acrshort*{cibs} algorithm is the RESOLVE\_NONCONCURRENCY procedure (Algorithm \ref{alg:resolve_concurrency}). It takes two blocks $b_i$ and $b_j$ of a PBD plan $\pi_{pbd}$ as input, where there is a necessary non-concurrency constraint, $b_i\#b_j$ and attempts to substitute block $b_i$ to remove $b_i\#b_j$. It first extends the block $b_i$ to $b_i'$ to encapsulate the additional necessary blocks for the substitution by calling the EXTEND procedure(presented in Algorithm \ref{alg:extend}). Since every block is a patial-order subplan, we design a subtask   $\Pi_{sub}$ to find suitable candidate subplans to substitute the block $b_i'$ under the condition that the substituting subplan has no non-concurrency constraint with $b_j$. 

When establishing the initial state $\hat{s}_0$ of $\Pi_{sub}$, we assume that the predecessors and successors of the candidate substituting block $\hat{b}_i$ will be the same as $b_i$. We create a subplan $\pi'$ by linearizing blocks, starting from the initial block, and advancing through the blocks preceding $b_i'$. Following that, $\hat{s}_0$ is set to the state resulting from applying $\pi'$ to the initial state of $\Pi$. We estimate the goal for $\Pi_{sub}$ by combining the facts (denoted as G) that $b_i'$ achieves for other blocks, and the facts (denoted as $C$) that $b_i'$'s predecessors achieve for its successors in $\pi_{pbd}$. It is important to note that the fact in $C$ also belongs to $\hat{s}_0$ (i.e., $C\subset \hat{s}_0$). Therefore, the solutions of $\Pi_{sub}$ do not necessarily produce the facts in $C$ but can not delete them. We include these facts in the goal so that the candidate block $\hat{b}_i$ does not delete the facts that its predecessor provides to its successors. Because in that scenario, $\hat{b}_i$ will pose threats to those causal links, and adding promotion or demotion ordering can not resolve these threats.
 We employ an off-the-self planner (c.g., the LAMA planner \cite{lama}) to generate multiple solutions $\hat{\pi}$ (i.e., subplans) for $\Pi_{sub}$ with a time-bound. We set the cost bound to the cost of $b_i$. We convert each solution  $\hat{\pi}$ to a parallel plan  $\hat{\pi}_{pop}$ by using \acrshort*{eog}, and then identifying non-concurrency constraints between each operator pair in $\hat{\pi}_{pop}$. Afterward, we create a  block $\hat{b}_i$ from $\hat{\pi}_{pop}$.  If $\hat{b}_i$ does not have a non-concurrency constraint with $b_j$ (i.e., $vars_{nc}(\hat{b}_i\#b_j) = \emptyset$), then $\hat{b}_i$ serves as a candidate subplan for substitution.  If substituting $b_i'$ with $\hat{b}_i$ yields success, then this procedure accepts the resultant PBD plan $\hat{\pi}_{pbd}$  under two conditions: 1) concurrency ($\mathit{cflex}$) of $\hat{\pi}_{pbd}$ exceeds that of $\pi_{pbd}$, and 2) the cost of $\hat{\pi}_{pbd}$ is not greater than that of $\pi_{pbd}$. Therefore, this procedure does not compromise the plan's cost while improving its concurrency.

\begin{theorem}
(Correctness of Resolve Nonconcurrency Algorithm) Given a valid PBD plan $\pi_{pbd}=\langle \mathcal{O},\mathcal{B} \prec, \#\rangle$ for a planning task $\Pi$, and two blocks $b_i, b_j \in \mathcal{B}$ with $b_i\#b_j$, if the algorithm returns true, then the resulting PBD plan is valid and the non-concurrency constraint between $b_i$ and $b_j$ is also resolved.
\end{theorem}

\begin{proofsketch}
We will prove that if the algorithm substitute $b_i$ with a subplan, encapsulated in a block $\hat{b}_i$ successfully, then i) there is no non-concurrency between $\hat{b}_i$ and $b_j$ in the updated PBD plan $\hat{\pi}_{pbd}=\langle \hat{\mathcal{O}}, \hat{\mathcal{B}}, \hat{\prec}, \hat{\#}\rangle$ with $\hat{\mathcal{B}}=(\mathcal{B} \setminus b_i)\cup \hat{b}_i$, and ii) $\hat{\pi}_{pbd}$ is valid w.r.t. $\Pi$.

 The algorithm proceeds through one of the following three scenarios. Situation (1):  no candidate block $\hat{b}_i$  
 is found such that $\hat{b}_i$ does not have any non-concurrency constraint with $b_j$ (i.e., $vars_{nc}(\hat{b}_i\# b_j)\ne \emptyset$), then the algorithm returns false along with the original PBD plan $\pi_{pbd}$, ensuring no incorrect modifications. Situation (2): If a candidate block $\hat{b}_i$  is found s.t. $vars_{nc}(\hat{b}_i\# b_j)= \emptyset$, but the SUBSTITUTE procedure  (Algorithm \ref{alg:block_substitute}) fails due to unresolved threats, and substitution is not performed. In this scenario, this algorithm also yields false with the original PBD plan $\pi_{pbd}$, preserving correctness. Situation (3): the SUBSTITUTE procedure successfully substitute $b_i$ with $\hat{b}_i$, the algorithm returns the updated PBD plan $\hat{\pi}_{pbd}$. Since the substitution is successful, the PBD plan returned by the SUBSTITUTE procedure is valid, as proven in Theorem \ref{theorem:substitution}. As $vars_{nc}(\hat{b}_i\# b_j)= \emptyset$, no new non-concurrency constraints are introduced in this procedure. Hence, the correctness of the algorithm is established.
\end{proofsketch}


\section{Experimental Result and Discussion}\label{sec:CIBS_exp}

One of the primary motivations for providing concurrency in a plan is to reduce its overall duration, while another important reason is to improve efficiency for parallel scheduling.   However, finding the minimum execution duration (Parallel Plan Length) in a parallel plan is a combinatorial problem (NP-hard), as demonstrated by Bäckström  \cite{backstorm1994}. Therefore, we introduce the metric $\mathit{cflex}$, which is the ratio of concurrent operator pairs to the total number of operator pairs. Given a parallel plan $\pi_{p}=\langle \mathcal{O}, \prec, \#\rangle$, 
    \begin{equation}
    \mathit{cflex}(\pi_{p}) = 1- \frac{\sum\limits_{1\le j<i\le |\mathcal{O}|}
    \begin{cases}
        1& \text{if }o_i \prec o_j \text{ or }  o_j \prec o_i \text{ or } o_i \# o_j\\
        0 & \text{otherwise}
    \end{cases}
    }{\Sigma_{i=1}^{|\mathcal{O}|-1}i}
    \end{equation} 

The denominator $\Sigma_{i=1}^{|\mathcal{O}|-1}i$, equivalent to $^{|\mathcal{O}|}C_2$, calculates the total number of pairs that can be formed from a set of  $|\mathcal{O}|$ elements. The value of \cflex{} ranges from 0 to 1. The higher \cflex{}, the more concurrency the POP has.

\begin{figure}
    \centering
    \includegraphics[width=0.8\linewidth]{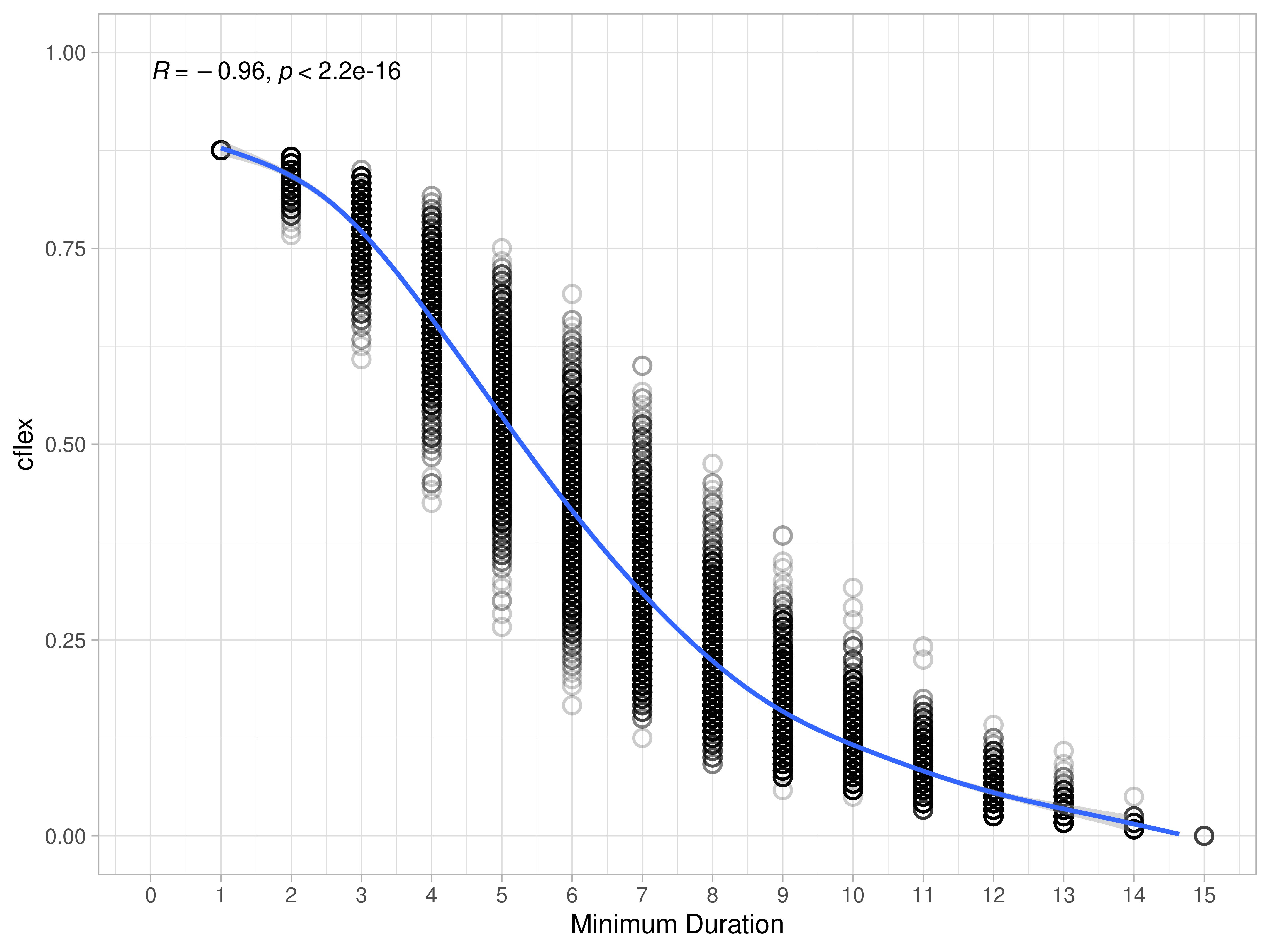}
    \caption{Comparison of the minimum execution duration and the $cflex$ value of approximately 10,000 random parallel plans with 15 actions. Every point represents a unique parallel plan with 15 actions.}
    \label{fig:cflex-just}
\end{figure}

On the other hand, the flexibility of executing actions in different sequences is calculated by the ratio of the number of unordered action pairs to the total number of action pairs and is referred to as \flex{} \cite{siddiqui_patrik_2012, maxsat}. Given a POP $\pi_{pop}=\langle \mathcal{O}, \prec\rangle$, 
    \begin{equation}
    \mathit{flex}(\pi_{pop}) = 1- \frac{|\prec|}{\Sigma_{i=1}^{|\mathcal{O}|-1}i}
    \end{equation} 
 \flex{} indicates the number of linearizations of a POP \cite{maxsat}, and its value ranges from 0 to 1.

To demonstrate the correlation between a POP's $\mathit{cflex}$ and minimum execution duration, we constructed $10,000$ random parallel plans ($7,052$ of them unique) with 15 operators (excluding the INIT and GOAL operators). We also randomly select the number of ordering and non-concurrency constraints for each plan.
Then, random pairs of operators are selected to add constraints between them. The minimum duration for each plan is calculated from linearizations of all possible (15!). The reason for using randomly generated plans is the need for a sufficient number of examples for the trend to become apparent. A comparison of plans with varying numbers of operators was uninformative. Figure \ref{fig:cflex-just}  shows \cflex{} of these random parallel plans as a function of the minimum duration, assuming unit time for each operator. The blue line representing the regression line in this plot visualizes the trend. The Spearman correlation coefficient between \cflex{} and the minimum duration is -0.96, indicating a strong negative correlation. Therefore, \cflex{} can act as a highly informative indicator of the minimum execution duration of a parallel plan; this implies that a higher \cflex{} value of a parallel plan corresponds to a lower minimum execution duration.

We have experimented the \acrshort*{cibs} algorithm (Algorithm \ref{alg:cibs}) with the problems of the international planning competition (IPC) domains from sequential satisfying tracks. We have excluded domains with conditional effects \cite{Nebel_2000} because the algorithms of this paper do not account for conditional effects. We use LAMA planner \cite{lama}, a two-time champion in international planning competitions, for rendering plans with a 30-minute time-bound for each problem. We employ an 8-core, 2.80GHz Core i7-1165G7 CPU to conduct all experiments with a 30-minute time limit. For comparative analysis, we use the Wilcoxon signed-rank test \cite{Wilcoxon1992}, which generates a z-score that is subsequently converted into a p-value. If the p-value is lower than 0.05, we can conclude that there is a significant difference between the outcomes of the two methods. We use asterisks(*) for the significance levels of p-values: one asterisk(*) for 0.05 to $> 0.01$, and two asterisks(**) for $< 0.01$.

\begin{table}[!tbp]
    \centering
    \caption{Experimental dataset. The number of problems and plans analyzed are provided in the \emph{Problems} and \emph{Plans} columns. The average numbers of plan operators, ground operators, variables, and independent variables are presented in \emph{Plan Size}, \emph{Ground Operators}, \emph{Vars}, and \emph{Independent Vars} columns, respectively.}
    \label{tbl:dataset}
    \begin{filecontents*}{domain_info1.csv}
domain,problems,plans,initial_size,goperators,variables,av
"barman(IPC7{,} IPC8)",20,54,206,2131,317,0
blocks(IPC2),50,148,77,466,29,0
child-snack(IPC8),6,8,68,3523,71,3
data-network(IPC9),12,17,131,6911,256,0
depots(IPC3),21,73,57,1818,43,2
"elevator(IPC2{,} IPC6{,} IPC7)",50,304,52,2407,18,2
"floor-tile(IPC7{,} IPC8)",7,44,47,215,21,2
"freecell(IPC2{,} IPC3)",50,191,55,13158,69,0
genome-edit-distances(IPC8),20,93,121,7592,46,0
grid(IPC1),5,15,65,8586,24,0
gripper(IPC1),20,20,70,188,26,1
hiking(IPC8),20,111,50,9283,14,0
"logistics(IPC1{,} IPC2)",50,161,86,6112,34,21
mystery(IPC1),19,28,11,9130,40,0
mystery-prime(IPC1),30,61,11,15397,40,0
no-mystery(IPC7),11,28,33,4784,11,0
parc-printer(IPC6),30,116,47,319,63,1
"parking(IPC7{,} IPC8)",20,144,79,67111,107,0
pathways(IPC5),22,36,147,1467,233,0
"peg-solitaire(IPC6{,} IPC7)",30,258,31,185,34,0
"pipesworld(IPC4{,} IPC5)",43,392,41,7572,265,0
"rovers(IPC3{,} IPC5)",40,143,76,3388,123,6
"satellite(IPC3{,} IPC4)",36,158,88,42891,72,6
scanalyzer-3d(IPC7),20,118,41,55827,24,0
storage(IPC5),20,54,26,1304,30,0
tetris(IPC8),19,25,65,35023,2308,0
thoughtful(IPC8),15,47,91,6723,261,0
tidybot(IPC7),17,58,62,39688,432,0
tpp(IPC5),30,58,120,7738,143,3
"transport(IPC6{,} IPC7{,} IPC8)",20,87,196,44987,22,3
trucks(IPC5),18,69,36,3814,20,0
"woodworking(IPC6{,} IPC7)",30,130,58,4811,232,0
zenotravel(IPC3),20,54,33,5218,16,0
Total,821,3303,65,13205,115,2
\end{filecontents*}
\pgfplotstabletypeset[
    column name={},
    column type=l,
    every head row/.style={
    before row={
    \toprule
    \multirow{2}{*}{Domains(Contests)} & \multirow{2}{*}{Problems} & \multirow{2}{*}{Plans} & \multirow{2}{*}{Plan Size} & Ground& \multirow{2}{*}{Vars} & Independent \\
    },
    after row=\midrule,
    },
    every last row/.style={before row=\toprule,after row=\midrule},
    columns/domain/.style ={column type/.add={@{\hspace{2pt}}}{}},
    columns/problems/.style ={column type=r, column type/.add={@{\hspace{0pt}}}{}},
    columns/plans/.style={column type=r, column type/.add={@{\hspace{9pt}}}{}},
    columns/goperators/.style={column name=Operators,column type=r, column type/.add={@{\hspace{9pt}}}{}},
    columns/initialsize/.style={column type=r, column type/.add={@{\hspace{9pt}}}{}},
    columns/variables/.style={column type=r, column type/.add={@{\hspace{9pt}}}{}},
    columns/{av}/.style={column name=Vars, column type=r, column type/.add={@{\hspace{7pt}}}{}},
    col sep=comma,
    string type,
    ignore chars={"}
     ]{domain_info1.csv}
\end{table}

Table \ref{tbl:dataset} presents our dataset comprising 3303 plans from 821 problems across 33 distinct domains. Besides the number of problems and plans, the table also provides the average number of plan sizes, ground operators, variables, and independent variables in the planning tasks for each domain. In a planning task, the independent variables are those whose transitions do not rely on other variables. For instance, in the Elevator domain, variables representing the locations of lifts are independent variables, as their transitions from one floor to another do not depend on the values of other state variables in this domain. Therefore, independent variables signify resources that are mainly responsible for non-concurrent constraints among subplans. The plan size, the number of ground operators, variables, and independent variables in a planning problem can have a significant impact on the performance of the algorithms presented in this work.

\begin{table}[!tp]
\caption{
Experimental results of the \acrshort*{cibs} algorithm. For each domain, columns \cflex{} and $T$ present mean \cflex{} and mean execution time in seconds for the phases (EOG, BD, and SC) of CIBS, respectively.  The asterisks denote the significance level for the p-values obtained from paired Wilcoxon signed-rank tests performed on the \cflex{} values of each phase relative to its preceding phase.}
    \label{tbl:cibs}
    \centering
    \begin{filecontents*}{concurrent_result.csv}
"domain","eogcf","eogtime","bdcf","bdtime","sdcf","t"
"barman","0.004","9.84","0.005","429.65","**0.006","476.84"
"blocks","0","0.47","0","36.62","0","45.65"
"child-snack","0.695","0.27","*0.721","3.64","0.725","5.46"
"data-network","0.282","1.35","0.296","61.49","*0.301","124.76"
"depots","0.265","0.44","**0.279","17.98","0.279","21.44"
"elevator","0.162","0.61","**0.167","27.04","0.168","41.25"
"floor-tile","0.147","0.06","*0.159","0.55","0.164","0.85"
"freecell","0.077","0.66","0.078","4.44","*0.079","5.54"
"genome-edit-distances","0.005","4.29","*0.006","82.17","0.006","167.41"
"grid","0","0.47","0","2.81","0","3.57"
"gripper","0.017","0.47","0.017","6.66","0.017","7.33"
"hiking","0.056","0.33","0.057","1.17","*0.058","1.41"
"logistics","0.576","0.91","**0.583","31.22","**0.585","44.9"
"mystery","0.123","0.15","0.123","0.16","0.123","0.16"
"mystery-prime","0.176","0.22","0.176","0.22","0.176","0.23"
"no-mystery","0.037","0.08","**0.042","0.22","0.042","0.82"
"parc-printer","0.629","0.1","0.629","0.22","0.629","0.33"
"parking","0.003","3.29","**0.003","6.94","*0.003","12.52"
"pathways","0.505","1.93","**0.54","236.1","0.54","320.95"
"peg-solitaire","0","0.03","0","0.45","0","1.12"
"pipesworld","0.167","0.91","**0.172","5.79","**0.174","6.55"
"rovers","0.705","0.47","0.708","28.36","**0.716","31.13"
"satellite","0.436","1.55","0.436","27.02","0.436","53.59"
"scanalyzer-3d","0.068","2.44","**0.068","3.26","0.07","3.48"
"storage","0.12","0.06","0.11","0.45","0.11","0.59"
"tetris","0.536","41.29","*0.542","42.68","*0.546","43.61"
"thoughtful","0.085","1.66","0.085","4.33","0.085","6.29"
"tidybot","0.067","7.58","0.067","12.65","0.067","14.61"
"tpp","0.315","2.09","**0.409","72.82","*0.417","122.87"
"transport","0.456","5.76","0.454","160.31","0.457","176.82"
"trucks","0.025","0.06","0.025","0.19","0.025","0.24"
"woodworking","0.907","0.66","0.907","0.96","0.907","1.21"
"zenotravel","0.389","0.1","**0.391","0.65","0.391","0.8"
"Total","0.237","1.61","**0.241","25.15","**0.242","34.67"
\end{filecontents*}
\pgfplotstabletypeset[
    column type=l,
    every head row/.style={
    before row={
    \toprule
    \multirow{2}{*}{Domains}  & \multicolumn{2}{c}{EOG} & \multicolumn{2}{c}{BD} & \multicolumn{2}{c}{SC}\\
    \cmidrule(lr){2-3}\cmidrule(lr){4-5}\cmidrule(lr){6-7}
    },
    after row=\midrule,
    },
    every last row/.style={after row=\bottomrule, before row =\midrule},
    display columns/0/.style={column type={p{.25\textwidth}}},
    columns/domain/.style ={column name=},
    columns/eogcf/.style ={column name=$\mathit{cflex}$, column type=r,  column type/.add={@{\hspace{20pt}}}{}},
     columns/eogtime/.style ={column name=$\mathit{T}$, column type=r,  column type/.add={@{\hspace{20pt}}}{}},
    columns/bdcf/.style ={column name=$\mathit{cflex}$, column type=r,  column type/.add={@{\hspace{20pt}}}{}},
    columns/bdtime/.style ={column name=$\mathit{T}$, column type=r,  column type/.add={@{\hspace{20pt}}}{}},
     columns/sdcf/.style ={column name=$\mathit{cflex}$, column type=r,  column type/.add={@{\hspace{20pt}}}{}},
      columns/t/.style ={column name=$T$, column type=r,  column type/.add={@{\hspace{20pt}}}{}},
    col sep=comma,
    string type,
     ignore chars={"},
    ]{concurrent_result.csv}
    \end{table}

\begin{table}[!tp]
\caption{Domain-wise, the number of plans with improvement in $\mathit{cflex}$ during the BD (Block Deordering) and SC (Substitution for Concurrency) phases of the CIBS algorithm. The column $BD_{EOG}$ shows the number of plans improved over EOG, while columns $SC_{BD}$ and $SC_{EOG}$ present the number of plans improved over the BD and EOG, respectively. The domains in which BD and SC phases could not improve $\mathit{cflex}$ of any plan are excluded. 
}
    \label{tbl:imp}
    \centering
    \begin{filecontents*}{improvement.csv}
domain,bd,sd,total
barman,18,15,22
child-snack,8,1,8
data-network,7,7,8
depots,41,1,41
elevator,95,2,95
floor-tile,28,4,29
freecell,38,14,47
genome-edit-distances,14,0,14
grid,1,0,1
hiking,30,7,34
logistics,71,29,72
no-mystery,15,0,15
parking,4,6,9
pathways,26,0,26
pipesworld,61,21,66
rovers,62,30,70
scanalyzer-3d,9,3,11
storage,14,1,14
tetris,9,8,10
thoughtful,1,0,1
tidybot,3,0,3
tpp,45,10,46
transport,18,2,20
trucks,6,0,6
zenotravel,19,0,19
Total,643,161,687
\end{filecontents*}
\pgfplotstabletypeset[
    column type=l,
    every head row/.style={
    before row={
    \toprule
    },
    after row=\midrule,
    },
    every last row/.style={after row=\bottomrule, before row =\midrule},
    display columns/0/.style={column type={p{.25\textwidth}}},
    columns/domain/.style ={column name=Domains},
    columns/bd/.style ={column name=$BD_{EOG}$, column type=r,  column type/.add={@{\hspace{22pt}}}{}},
     columns/sd/.style ={column name=$SC_{BD}$, column type=r,  column type/.add={@{\hspace{22pt}}}{}},
    columns/total/.style ={column name=$SC_{EOG}$, column type=r,  column type/.add={@{\hspace{22pt}}}{}},
    col sep=comma,
    string type,
     ignore chars={"},
    ]{improvement.csv}
    \end{table}
Table \ref{tbl:cibs} presents the experimental results of the \acrshort*{cibs} for each domain. We estimate $\mathit{cflex}$ after performing each of the three phases in \acrshort*{cibs}: 1) \acrfull*{eog}, 2) \acrfull*{bd}, and 3) Substitution for Concurrency (SC). The table also provides the number of plans with improved \cflex{} in each phase in parentheses. 
    
The results demonstrate a general trend of increasing \cflex{} across phases, with notable variations across different domains. The mean \cflex{} improves from 0.237 in EOG to 0.241 in BD and 0.242 in SC. However, these improvements come at a computational cost, as execution time rises significantly from 1.61s in EOG to 25.15s in BD and 34.67s in SC. Table \ref{tbl:imp} provides the number of plans improved in the BD and SC phases. \acrshort*{bd} and SC phases individually remove non-concurrency constraints in 643 and 161 plans, respectively. The total number of plans with improved \cflex{}, is 643 plans in BD and 687 in SC, compared to EOG. 

According to the results of the Wilcoxon signed-rank test, BD improved $\mathit{cflex}$ in a significant number of plans in 14 domains, such as child-snack, depots, elevator, floor-tile, logistics. Previous works 
\cite{siddiqui_patrik_2012, sabah_siddiqui_2022, MaxSaT_sabah} demonstrate that block deordering significantly enhances the value of plan flexibility compared to EOG. However,  as illustrated in Example \ref{ex:example1}, the non-concurrency constraint often persists between a block pair, even though block deordering effectively eliminates their ordering constraints. That is why, the impact of block deordering on $\mathit{cflex}$ is not as substantial as its effect on $\mathit{flex}$. On the other hand, SC can only improve concurrency through substitution when resources (e.g., elevators, robots, cars, etc.) are available in the corresponding problem task.  In our experiment, SC improved concurrency in a significant number of plans in 10 domains, specifically barman, data-network, freecell, hiking, logistics, parking, pipesworld, rovers, tetris, and tpp. The overall improvement appears modest because, in many domains, concurrency cannot be enhanced beyond EOG due to the inherent structure of the domain or the unavailability of alternative resources within the corresponding problem.

\begin{sidewaysfigure}[!tbp]
    \centering
    \includegraphics[width=\linewidth]{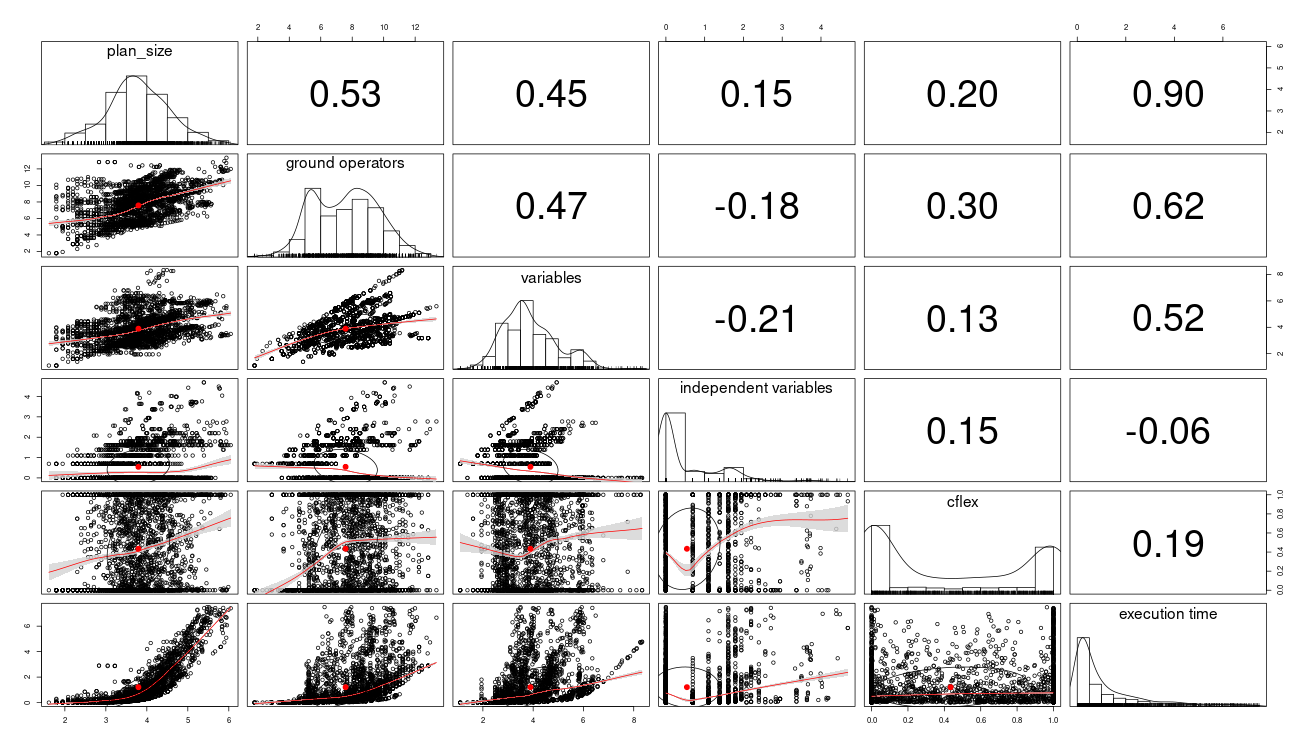}
    \caption{ Correlation analysis among \cflex{}, execution time after performing \acrshort*{cibs}, plan size, the number of ground operators, state variables, and independent variables. The distribution of each variable is displayed on the diagonal. Below the diagonal, the bivariate scatter plots with a fitted line are shown, and the correlation values are provided above the diagonal.}
    \label{fig:cibs_cor}
\end{sidewaysfigure}

We also investigate the correlation among various variables, which include \cflex{}, execution time after performing \acrshort*{cibs}, plan size, the number of ground operators, state variables, and independent variables.  We apply the Spearman correlation coefficient \cite{spearman_correlation} to measure the correlation since it uses ranks instead of assumptions of normality. Spearman correlation determines the strength and direction of the monotonic relationship between two variables. A monotonic relationship is defined by a consistent directional change between the two variables: as one variable increases (or decreases), the other consistently increases (or decreases).
A coefficient ranging from 0 to 0.3 (-0.3 to 0) suggests a weak association, while 0.4 to 0.6 (-0.4 to -0.6) indicates moderate strength. A coefficient between 0.7 and 1 (-0.7 to -1) signifies a strong relationship. Figure \ref{fig:cibs_cor} presents the correlation analysis  of these variables.  In this figure, the distribution of each variable is displayed on the diagonal. Below the diagonal, the bivariate scatter plots with a fitted line are shown, and the correlation values are provided above the diagonal.  This figure illustrates that the execution time of \acrshort*{cibs} is strongly correlated with plan size, and moderately correlated with the number of ground operators and the number of variables in the corresponding planning task. There is no significant correlation among the other variables.

\begin{figure}[!tbp]
    \centering
    \includegraphics[width=.85\linewidth]{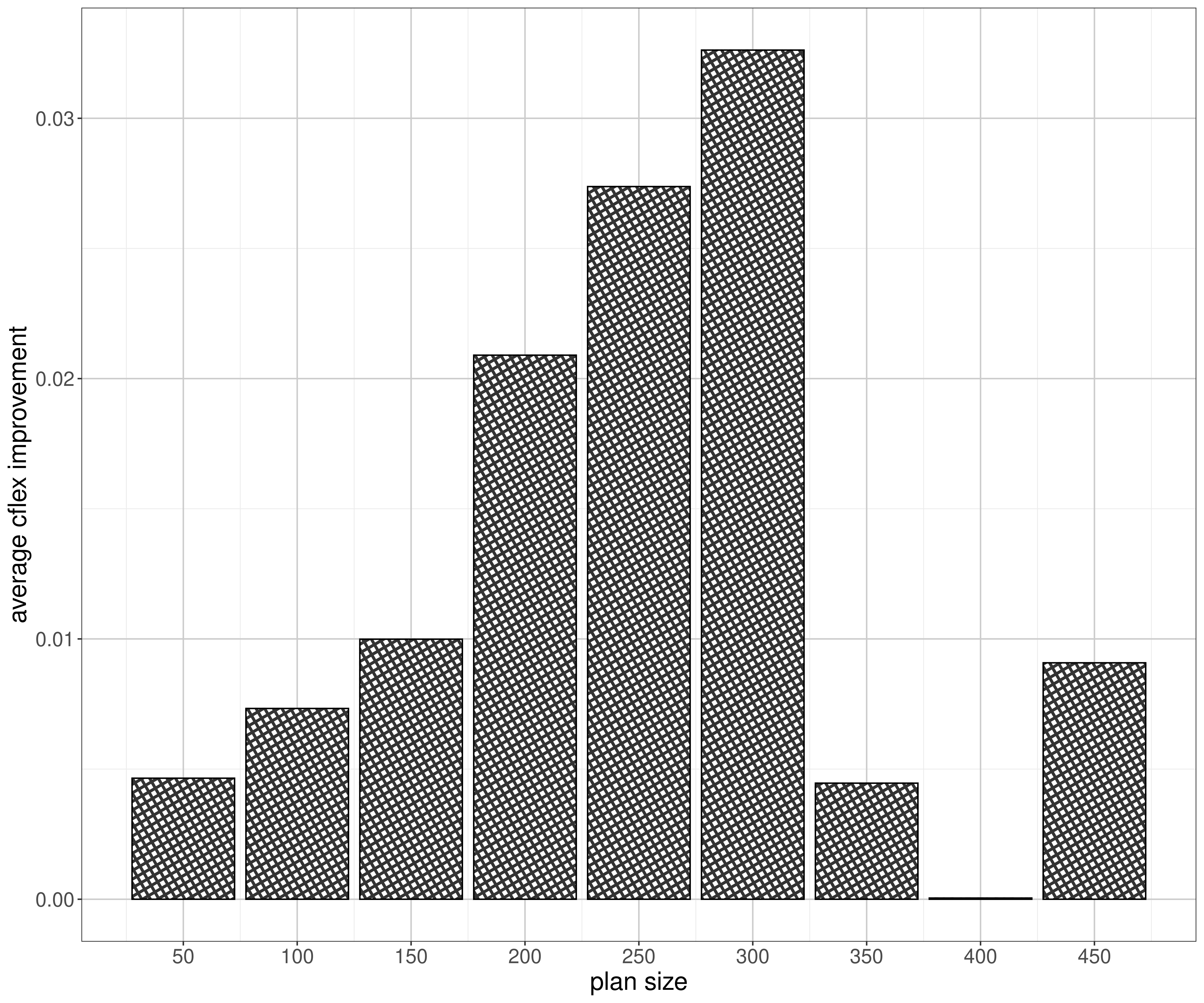}
    \caption{Mean $\mathit{cflex}$ improvement after performing \acrshort*{cibs} as a function of plan size. }
    \label{fig:cibs_size}
\end{figure}

\begin{figure}[!tbp]
    \centering
    \includegraphics[width=.85\linewidth]{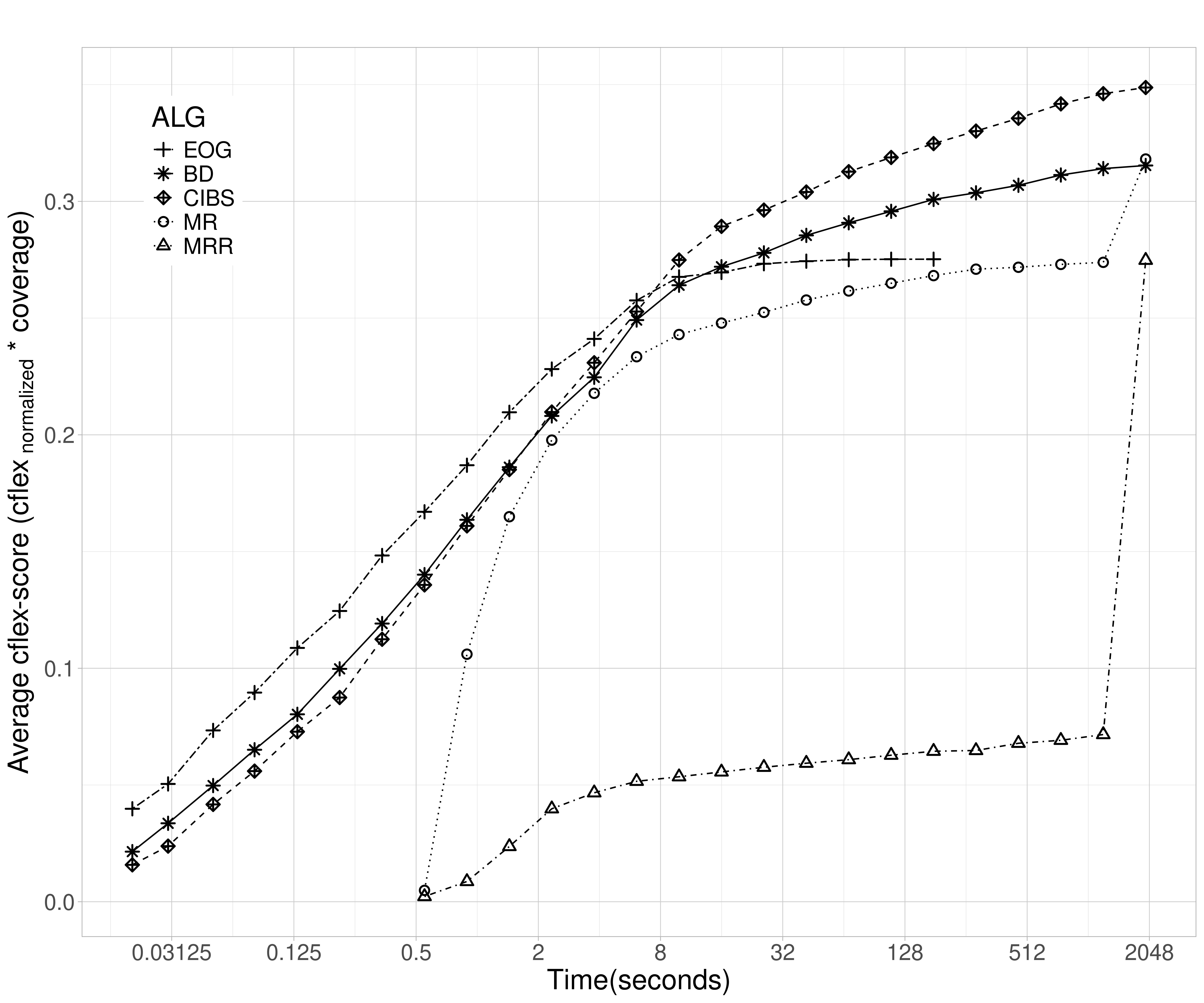}
    \caption{Average $\mathit{cflex}{-}score$ as a function of time after performing EOG, BD, \acrshort*{cibs}, and MaxSAT reorderings (MR and MRR) for 3265 plans from \acrshort*{ipc} domains. $\mathit{cflex}$ is normalized to the interval between the lowest and highest $\mathit{cflex}$ found for a problem. The x-axis is displayed on a log scale. The spikes in MR and MRR performance curves are due to their satisfiable solutions provided by the end of the time limit.}
    \label{fig:line_plot_concurrent}
\end{figure}

Figure \ref{fig:cibs_size} illustrates the mean \cflex{} improvement by CIBS across plan sizes ranging from 1 to 450 at intervals of 50. The average \cflex{} improvement increases as the plan size grows from 50 to 300. The highest improvement is observed at a plan size of approximately 300, reaching above 0.03. This indicates that larger plans (up to 300 operators) benefit more from applying CIBS. However, beyond the size of 350, there's a sudden and significant drop in $cflex$ improvement for larger plans, suggesting that the algorithm effectively improves execution flexibility for medium-sized plans (50-300 operators) but has limitations in scalability when applied to very large plans (350+ operators). This is primarily due to two reasons. First, the correlation plot in Figure 11 clearly shows that plan size and execution time are strongly correlated. Our method CIBS cannot substantially improve the concurrency of plans around the size of 350 within the time limit of 1800 seconds. The second reason is that the number of plans around size 350 is limited in number, mainly from the domains blocks, satellites, rovers, elevators, and transport, totaling 10 to 15 plans in each interval from 300 to 450. This is also evident in the histogram of plan size, shown at the top of Figure 11. Thus, the opportunity for improving flex in this range is minimal as well.

We have also conducted performance analysis among iterative approaches and MaxSAT-based approaches. The MaxSAT encodings by Muise et al. \cite{maxsat} to transform POPs into MaxSAT instances for minimizing orderings of a POP  is known as \emph{MR} (Minimum Reordering). Building upon this foundation, Waters et al. further enhance the encoding by incorporating formulas that enable the reinstantiation of operator parameters. This extended encoding is termed \emph{MRR} (Minimal Reinstantiated Reordering). Figure \ref{fig:line_plot_concurrent}  presents a comparative performance analysis of five algorithms (EOG, BD, CIBS, MR, and MRR) based on the \cflex{} and coverage over execution time in seconds. Here, to clearly distinguish their performances, we normalized the \cflex{} value to the interval between the lowest and
highest cflex found for a problem.  The $\mathit{cflex}{-}score$ is estimated
by multiplying the average normalized flex by the coverage within a specified time
frame. Here, coverage represents the proportion of plans for which the respective
method has completed computation out of the total number of plans. 

The iterative approaches (EOG, BD and CIBS) start providing solutions in less than 0.03 seconds, while MR and MRR take nearly 0.5 seconds to start. EOG’s performance levels off after approximately 32 seconds, achieving an average $\mathit{cflex}{-}score$ close to 0.27. EOG provides solutions for all plans within 128 seconds. BD and CIBS intersect EOG’s performance line around 8 seconds. \acrshort*{cibs} follows a trajectory similar to block deordering but surpasses it after 2 seconds, consistently maintaining its position above the block deordering line, indicating that as BD generates additional blocks for deordering, CIBS gains more chances to improve concurrency by replacing blocks.

The early progress of MR is comparable to that of other algorithms. Eventually, it slows down significantly, reflecting inefficiencies in handling larger problems. MRR performs poorly due to its high computational cost and lowest coverage. It may yield better results if given extensive computational time.  When MR and MRR fail to find an optimal solution for a plan within a given time limit, they provide a satisfiable solution if one is found. While 87\% of MR reorderings are optimal, MRR achieves optimality in only 38\% of its solutions. As a result, the remaining 13\% of MR solutions and 62\% of MRR solutions are satisfiable, and these are provided at the end of the time limit. This explains why MRR exhibits a more pronounced spike in performance near the time limit compared to MR.

Notably, it is also permissible to take the MaxSAT approaches as a base method to convert the given sequential plans into POPs, and then apply block deordering and block substitution to enhance concurrency. We select EOG due to its ability to compute partial orderings in polynomial time.

One limitation of our approach is that CIBS does not explore other subplans beyond those enclosed in blocks for possible substitution. Block deordering does not create blocks that do not contribute to eliminating any ordering in the plan. Consequently, there are cohesive operator sets that may not be captured by block deordering. Therefore, there is a future scope in devising methods to decompose a plan into a BDPO plan that captures cohesive subplans responsible for different subtasks of the corresponding problem, without concerning the plan flexibility. This can be accomplished by utilizing DTG in FDR, similar to the method outlined in Section \ref{sec:extend} to extend a block to its cohesive neighbors.

\section{Conclusion}
This study establishes the necessary and sufficient conditions for a non-concurrency constraint between operators or blocks of operator (i.e., subplans) for a planning problem under \acrshort*{fdr} encodings, allowing the transformation of a \acrshort*{pop} into a parallel plan. Then, we enhance the concurrency of the parallel plan by employing block deordering and block-substitution in our algorithm \acrshort*{cibs}. Our experimental analysis demonstrates that CIBS significantly enhances the concurrency of plans from \acrshort*{ipc} domains.

\acrshort*{cibs} exploits PBD plans by using blocks as candidate subplans for substitution instead of blindly searching for suitable subplans, resulting in lower computational costs. However, our algorithm does not explore other subplans not enclosed by blocks within a PBD plan. We can explore alternative approaches, such as forming random blocks, to identify candidate subplans for substitution. Additionally, we can extend the application of block deordering and block substitution in the class of parallel plans with interacting operators.

\FloatBarrier
\bibliography{bibfile}
\end{document}